\documentclass[acmtocl,acmnow]{acmtrans2m}

\usepackage[arrow,matrix]{xy}
\usepackage{myStyle}                  %
\usepackage{array}
\usepackage{amsmath,amssymb}

\newtheorem{theorem}{Theorem}[section]

\newtheorem{proposition}[theorem]{Proposition}
\newtheorem{lemma}[theorem]{Lemma}
\newdef{definition}[theorem]{Definition}
\newdef{example}[theorem]{Example}
\renewcommand{\qed}{\hfill$\Diamond$}
\newcommand{\TRonly}[1]{}
\newcommand{\TOCLonly}[1]{#1}
\newcommand{\TrTocl}[2]{#2}

\title{Superposition for Fixed Domains}
\author{MATTHIAS HORBACH \and CHRISTOPH WEIDENBACH
  \\Max Planck Institute for Informatics
  \\Saarbr\"ucken, Germany
}
\date{\today}
\begin{abstract}
Superposition is an established decision procedure for a variety of first-order logic theories represented by sets of clauses. A satisfiable theory, saturated by superposition, implicitly defines a minimal term-generated model for the theory.
Proving universal properties with respect to a saturated theory directly leads to a modification of the minimal model's term-generated domain, as new Skolem functions are introduced. For many applications, this is not desired.

Therefore, we propose the first superposition calculus that can explicitly represent existentially quantified variables and can thus compute with respect to a given domain. This calculus is sound and refutationally complete in the limit for a first-order fixed domain semantics.
For saturated Horn theories and classes of positive formulas, we can even employ the calculus to prove properties of the minimal model itself, going beyond the scope of known superposition-based approaches.
\end{abstract}



\begin{document}

\begin{bottomstuff}
Authors' address: M.~Horbach and C.~Weidenbach, Automation of Logic Group,
Max Planck Institute for Informatics, Campus~E1~4, 66123 Saarbr\"ucken, Germany.
\end{bottomstuff}

\maketitle

\section{Introduction}

A formula $\Phi$ is entailed by a clause set $N$ with respect to the standard first-order semantics, written $N \models \Phi$, if $\Phi$ holds in all models of $N$ over all possible domains. For a number of applications, this semantics is not sufficient to prove all properties of interest. In some cases, properties with respect to models over the fixed given domain of $N$ are required.
These models are isomorphic to Herbrand models of $N$ over the signature $\Signature$, i.e.~models whose domains consist only of terms build over $\Signature$. We denote this by $N \models_{\Signature}\Phi$.
Even stronger, the validity of $\Phi$ often needs to be considered with respect to a minimal model $\perfect{N}$ of the clause set $N$, written $\perfect{N}\models \Phi$ or alternatively $N \models_{\Ind}\Phi$.
For the sets of formulas that are valid with respect to these three different semantics, the following relations hold: $\{\Phi \mid N \models_{\Ind} \Phi\}\supseteq \{\Phi \mid N \models_{\Signature}\Phi\} \supseteq \{\Phi \mid N \models\Phi\}$.

\begin{figure}
\begin{center}
\renewcommand{\arraystretch}{1.5}
\begin{tabular}{cc}
\begin{tabular}{l} 
  \\
  \\
 Restaurant\\
 Company\\
 Ground \\
 \\
\end{tabular} & 
\begin{tabular}{|cc|} 
\multicolumn{2}{l}{\hspace*{10ex}\setlength{\unitlength}{1ex}\begin{picture}(10,2)
\put(-11.5,0){\line(5,1){11.5}}
\put(0,2.3){\line(5,-1){11.5}}
\put(-11.5,0){\line(1,0){23}}
\end{picture}}\\ \hline
 & \\
\begin{tabular}{|p{3ex}|} \hline
\ \\ \hline
\ \\ \hline
\ \\ \hline
\end{tabular} &
\begin{tabular}{|p{3ex}|} \hline
\ \\ \hline
\multicolumn{1}{c}{\ } \\ \hline
\ \\ \hline
\end{tabular}\\
$a$ & $b$\\ \hline
\end{tabular}
\end{tabular}
\end{center}

\vspace*{-1em}  
\caption{Elevator Example}\label{figelev}
\end{figure}
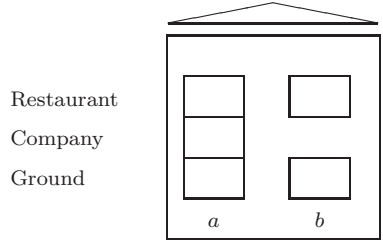

The different semantics are of relevance, for example, in proving properties of computer systems. Very often, such systems can be naturally modeled by first-order formulas over a fixed domain. Consider the simple example of a building with three floors, Figure~\ref{figelev}.

The bottom ($G$)round floor and the top ($R$)estaurant floor of the building are open to the public whereas the middle floor is occupied by a ($C$)ompany and only open to its employees. In order to support this setting, there are two elevators $a$ and $b$ in the building.
Elevator $a$ is for the employees of the company and stops on all three floors whereas elevator $b$ is for visitors of the restaurant, stopping solely on the ground and restaurant floor. Initially, there is a person $p$ in elevator $a$ and a person $q$ in elevator $b$, both on the ground floor.
  We model the system by three predicates $G$, $C$, $R$ for the different floors, respectively, where, e.g., $G(a,p)$ means that person $p$ sits in elevator $a$ on the ground floor.
  The initial state of the system and the potential upward moves are modeled by the following clauses:
$N_E = \{\to G(a,p)$, $\to G(b,q)$, $G(a,x)\rightarrow C(a,x)$, $C(a,x)\rightarrow R(a,x)$, $G(b,x)\rightarrow R(b,x)\}$.
Let us assume that the above predicates accept in their first argument elevators and in their second persons, e.g.~implemented via a many-sorted discipline.

The intended semantics of the elevator system coincides with the minimal model of $N_E$. 
Therefore, in order to prove properties of the system, we need to consider the semantics $\models_{\Ind}$ in general. 
Nevertheless, some structural properties are valid with respect to $\models$, for example the property that whenever a person (not necessarily $p$ or $q$) sits on the ground floor in elevator $a$ or $b$, they can reach the restaurant floor, i.e.~$N_E \models \forall x.(G(a,x)\rightarrow R(a,x)) \wedge (G(b,x) \rightarrow R(b,x))$.
In order to prove properties with respect to the specific domain of the system, we need to consider
$\models_{\Signature}$, for our example $\models_{\{a,b,p,q\}}$.
With respect to this semantics, the state $R$ is reachable for all elevators, i.e.\ 
$N_E \models_{\{a,b,p,q\}} \forall y,x.G(y,x)\rightarrow R(y,x)$. 
This property is not valid for $\models$ as there are models of $N_E$ with more elevators
than just $a$ and $b$. For example, there could be an elevator for the managers of the company that does not stop at the restaurant floor. Of course, such artificially extended models are not desired for analyzing the scenario.
For the above elevator system, the company floor is not reachable by elevator $b$. This can only be proven with respect to $\models_{\Ind}$, i.e.\ 
$N_E \models_{\Ind} \forall x. \neg C(b,x)$, but not with respect to $\models_{\{a,b,p,q\}}$ nor $\models$ because there are models of $N_E$ over $\{a,b,p,q\}$ where e.g.~$C(b,q)$ holds. %

In this simple example, all appearing function symbols are constants and the Herbrand universe is finite. Hence we could code the quantification over the Herbrand universe explicitly as
$\forall y,x.(y\eq a\vee y\eq b)\wedge (x\eq p\vee x\eq q)\wedge G(y,x)\rightarrow R(y,x)$. A property extended in this way is valid in all models of $N_E$ if and only if it is valid in all Herbrand models of $N_E$, i.e.~fixed domain reasoning can be reduced to first-order reasoning in this case. This reduction is, however, not possible when the Herbrand universe is infinite.

\smallskip
Inductive ($\models_\Ind$) and fixed-domain ($\models_\Signature$) theorem proving are more difficult problems than first-order ($\models$) theorem proving: It follows from G\"odels incompleteness theorem that inductive validity is not semi-decidable, and the same holds for fixed-domain validity.%
\footnote{In fact, peano arithmetic can be encoded in a fixed-domain setting as follows: Given the signature $\Signature_{\text{PA}}=\{s,0,{+},{\cdot}\}$, let $N$ consist of the clauses $x+0\eq 0$ and $x+s(y)\eq s(x+y)$ defining addition, $x\cdot 0\eq 0$ and $x\cdot s(y)\eq (x\cdot y)+x$ defining multiplication, and $s(x)\noteq 0$ and $s(x)\eq s(y)\to x\eq y$ stating that all numbers are different. Then $N$ has exactly one Herbrand model over $\Signature_{\text{PA}}$, and this model is isomorphic to the natural numbers. So an arithmetic formula $\Phi$ is valid over the natural numbers if and only if $N\models_\Ind\Phi$ if and only if $N\models_{\Signature_{\text{PA}}}\Phi$.}
  For the standard first-order semantics $\models$, one of the most successful calculi is superposition~\cite{BachmairGanzinger94,NieuwenhuisRubio01handbook,Weidenbach01handbook}.
This is in particular demonstrated by superposition instances effectively deciding many known decidable classical subclasses of first-order logic, e.g.\ the monadic class with equality~\cite{BachmairGanzingerEtAl93a} or the guarded fragment with equality~\cite{GanzingerNeville99}, as well as a number of decidable first-order classes that have been proven decidable for the first time by means of the superposition calculus~\cite{Nieuwenhuis96,JacquemardMeyerEtAl98,Weidenbach99cade,JacquemardRV06}.
Furthermore, superposition has been successfully applied to decision problems from the area of description logics~\cite{HustadtSchmidtGeorgieva04} and data structures~\cite{ArmandoBRS09}.
The key to this success is an inherent redundancy notion based on the term-generated minimal interpretation $\perfect{N}$ of a clause set $N$, that  restricts the necessary inferences and thereby often enables termination. If all inferences from a clause set $N$ are redundant (then $N$ is called \emph{saturated}) and $N$ does not contain the empty clause, then $\perfect{N}$ is a minimal model of $N$.

\smallskip
Consider the following small example, demonstrating again the differences of the three semantics with respect to the minimal term generated model induced by the superposition calculus.
The clause set  $N_G = \{\,\to G(s(0),0),\ G(x,y) \to G(s(x),s(y))\,\}$ is finitely saturated by superposition. 
The model $\perfect{N_G}$ in this example consists of all atoms $G(t_1,t_2)$ where $t_2$ is a term over the signature $\Signature_{\text{nat}}=\{s,0\}$ and $t_1=s(t_2)$. 
So the domain of $\perfect{N_G}$ is isomorphic to the naturals and the interpretation of $G$ in $\perfect{N_G}$ is the ``one greater than'' relation.  Now for the different entailment relations, the following holds:
\[
\begin{array}{l@{\quad}l@{\quad}l}
N_G \models G(s(s(0)),s(0))
   & N_G \models_{\Signature_{\text{nat}}} G(s(s(0)),s(0))
      & N_G \models_{\Ind} G(s(s(0)),s(0))\\
N_G \not\models \forall x. G(s(x),x)
   & N_G \models_{\Signature_{\text{nat}}} \forall x. G(s(x),x)
      & N_G \models_{\Ind} \forall x. G(s(x),x)\\
N_G \not\models \forall x. \neg G(x,x)
   & N_G \not \models_{\Signature_{\text{nat}}} \forall x. \neg G(x,x)
      & N_G \models_{\Ind} \forall x. \neg G(x,x)\\
\end{array}
\]

Superposition is a sound and refutationally complete calculus for the standard semantics $\models$.
In this paper, we develop a sound and refutationally complete calculus for $\models_\Signature$. 
Given a clause set $N$ and a purely existentially quantified conjecture, standard superposition is also complete for $\models_\Signature$. The problem arises with universally quantified conjectures that become existentially quantified after negation.
  Then, as soon as these existentially quantified variables are Skolemized, the standard superposition calculus applied afterwards no longer computes modulo $\models_\Signature$, but modulo $\models_{\Signature\cup\{f_1,\ldots,f_n\}}$, where $f_1,\ldots,f_n$ are the introduced Skolem functions.
This approach is incomplete:
  In the example above, $N_G \models_{\Signature_{\text{nat}}} \forall x. G(s(x),x)$, but the ground clause $G(s(c),c)$ does not hold in $\perfect{N_G}$, where $c$ is the Skolem constant introduced for $x$.

\smallskip
The idea behind our new calculus is not to Skolemize existentially quantified variables, but to treat them explicitly by the calculus. 
This is represented by an extended clause notion, containing a constraint for the existentially quantified variables. For example, the above conjecture $\forall x. G(s(x),x)$ results after negation in the clause $\clause{G(s(x),x)\to}{u\eq x}$ with existential variable $u$. 
In addition to standard first-order equational reasoning, the inference and reduction rules of  the new calculus also take care of the constraint (see Section~\ref{secforifx}).

A $\models_\Signature$ unsatisfiability proof of a constrained clause set with our calculus in general requires the computation of infinitely many empty clauses, i.e.~we lose compactness. This does not come as a surprise because we have to show that an existentially quantified clause cannot be satisfied by a term-generated infinite domain.
For example, proving the unsatisfiability of the set
$N_G\cup \{\clause{G(s(x),x)\to}{u\eq x}\}$
over the signature $\Signature = \{0,s\}$ amounts to the successive derivation of the clauses $\clause\Box{u\eq 0}$, $\clause\Box{u\eq s(0)}$, $\clause\Box{u\eq s(s(0))}$, and so on.
  In order to represent such an infinite set of empty clauses finitely, a further induction rule, based on the minimal model semantics $\models_{\Ind}$, can be employed.
We prove the new rule sound in Section~\ref{sec: fixdom and minmod validity} and show its potential.

In general, our calculus can cope with (conjecture) formulas of the form $\forall^*\exists^* \Phi$ and does not impose special conditions on $N$ (except saturation for $\models_{\Ind}$), which is beyond any known result on superposition-based calculi proving properties of $\models_\Signature$ or $\models_{\Ind}$~\cite{Kapur91,CaferraZabel92,GanzingerStuber92,Bouhoula97,ComonNieuwenhuis00,KapurSubramaniam00,GieslKapur03,Peltier03,FalkeKapur06}. 
This, together with potential extensions and directions of research, is discussed in the final Section~\ref{secrfrc}.

This article is a significantly extended version of \cite{HorbachWeidenbach08}.

\section{Preliminaries}
\label{sec:preliminaries}

We build on the notions of \cite{BachmairGanzinger94,Weidenbach01handbook} and shortly recall here the most important concepts as well as the specific extensions needed for the new superposition calculus.

\subsection*{Terms and Clauses}
Let $\Signature$ be a \emph{signature}, i.e.\ a set of function symbols of fixed arity, and $X\cup V$ an infinite set of variables, such that $X$, $V$ and $\Signature$ are disjoint and $V$ is finite. Elements of $X$ are called \emph{universal variables} and denoted as $x,y,z$, and elements of $V$ are called \emph{existential variables} and denoted as $u,v$. We denote by $\Terms(\Signature,X')$ the set of all \emph{terms} over $\Signature$ and $X'\subseteq X\cup V$ and by $\Terms(\Signature)$ the set of all \emph{ground terms} over $\Signature$. For technical reasons, we assume that there is at least one ground term, i.e.~that $\Signature$ contains at least one function symbol of arity $0$.

We will define equations and clauses in terms of multisets. A multiset over a set $S$ is a function $M\In S\to\Nat$. We use a set-like notation to describe multisets, e.g.\ $\{x,x,x\}$ denotes the multiset $M$ where $M(x)=3$ and $M(y)=0$ for all $y\not=x$ in $S$.
  An equation is a multiset $\{s,t\}$ of two terms, usually written as $s\eq t$.
  A (standard universal) \emph{clause} is a pair of multisets of equations, written $\Gamma\to\Delta$, interpreted as the conjunction of all equations in the \emph{antecedent} $\Gamma$ implying the disjunction of all equations in the \emph{succedent} $\Delta$. A clause is \emph{Horn} if $\Delta$ contains at most one equation. The \emph{empty clause} is denoted by $\Box$.

We denote the subterm of a term $t$ at \emph{position} $p$ by $t|_p$. 
The term that arises from $t$ by replacing the subterm at position $p$ by the term $r$ is $t[r]_p$.
  A \emph{substitution} $\sigma$ is a map from a finite set $X'\subseteq X\cup V$ of variables to $\Terms(\Signature,X)$, and $\dom(\sigma)=X'$ is called its \emph{domain}.\footnote{This notion of a domain is non-standard: Usually, the domain of a substitution is the set of all variables on which the substitution operates non-trivially. However, we want to be able to distinguish between substitutions like $\{x\mapsto f(x)\}$ and $\{x\mapsto f(x);\ y\mapsto y\}$ to simplify the proofs in Section~\ref{sec: fixdom and minmod validity}.}
The substitution $\sigma$ is identified with its homomorphic extension to $\Terms(\Signature,X\cup V)$.
The most general unifier of two terms $s,t\in\Terms(\Signature,X)$ is denoted by $\mgu(s,t)$.

\subsection*{Constrained Clauses}
A \emph{constrained clause} $\clause{C}{v_1\eq t_1,\ldots,v_n\eq t_n}$ consists of a conjunctively interpreted sequence of equations $v_1\eq t_1,\ldots,v_n\eq t_n$, called the \emph{constraint}, and a clause $C$, such that
\begin{enumerate}
\item $V=\{v_1,\ldots,v_n\}$,
\item $v_i\neq v_j$ for $i\neq j$, and 
\item neither the clause $C$ nor the terms $t_1,\ldots,t_n$ contain existential variables.
\end{enumerate}
Intuitively, constraint equations are just a different type of antecedent literals.
The constrained clause is called \emph{ground} if $C$ and $t_1,\ldots,t_n$ are ground, i.e.\ if it does not contain any non-existential variables.
  A constraint $\alpha=v_1\eq t_1,\ldots,v_n\eq t_n$ induces a substitution $V\to\Terms(\Signature,X)$ mapping $v_i$ to $t_i$ for all $i$, which we will denote by $\sigma_\alpha$.

Constrained clauses are considered equal up to renaming of non-existen\-tial variables. For example, \TrTocl{the two}{the} constrained clauses $\clause{P(x)}{u\eq x, v\eq y}$ and $\clause{P(y)}{u\eq y, v\eq x}$ are considered equal ($x$ and $y$ have been exchanged), but they are both different from the constrained clause $\clause{P(x)}{u\eq y, v\eq x}$, where $u$ and $v$ have been exchanged.
We regularly omit constraint equations of the form $v_i\eq x$, where $x$ is a variable, if $x$ does not appear elsewhere in the constrained clause, e.g.\  when $V=\{u,v\}$, we write $\clause{P(x)}{u\eq x}$ for $\clause{P(x)}{u\eq x, v\eq y}$.
A constrained clause $\clause C{}$ is called \emph{unconstrained}.
  As constraints are ordered, the notion of positions lift naturally to constraints.

\subsection*{Clause Orderings}

One of the strengths of superposition relies on the fact that only inferences involving maximal literals in a clause have to be considered, and that the conclusion of an inference is always smaller than the maximal premise. To state such ordering conditions, we extend a given ordering on terms to literal occurrences inside a clause, and to clauses.

Any ordering $\prec$ on a set $S$ can be extended to an ordering on multisets over $S$ as follows:
$M\prec N$ if $M\not=N$ and whenever there is $x\in S$ such that $N(x)<M(x)$ then $N(y)>M(y)$ for some $y\succ x$.

Considering this, any ordering $\prec$ on terms can be extended to clauses in 
the following way. 
We consider clauses as multisets of occurrences of equations. The occurrence of an equation $s\eq t$ in the antecedent is identified with the multiset $\{\{s,t\}\}$; the occurrence of an equation $s\eq t$ in the succedent is identified 
with the multiset $\{\{s\},\{t\}\}$. Now we lift $\prec$ to equation occurrences as its twofold multiset extension, and to clauses as the multiset extension of this ordering on equation occurrences. If, for example, $s\prec t\prec u$, then the equation occurrences in the clause $s\eq t, t\eq t\to s\eq u$ are ordered as $s\eq t\prec t\eq t\prec s\eq u$, because $\{\{s,t\}\}\prec\{\{t,t\}\}\prec\{\{s\},\{u\}\}$.
Observe that an occurrence of an equation $s\eq t$ in the antecedent is strictly bigger than an occurrence of the same equation in the succedent, because $\{\{s\},\{t\}\}\prec\{\{s,t\}\}$.

An %
occurrence of an equation $s\eq t$ is \emph{maximal} in a clause $C$ if there is no occurrence of an equation in $C$ that is strictly greater with respect to $\prec$ than the occurrence $s\eq t$. It is \emph{strictly maximal} in $C$ if there is no occurrence of an equation in $C$ that is greater than or equal to the occurrence $s\eq t$ with respect to $\prec$.

\TrTocl{W}{Moreover, w}e extend $\prec$ to constraints pointwise%
\footnote{
  It is also possible to consider constraints as multisets when ordering them, or to extend
  the ordering lexicographically. While all results of this article remain valid in both cases,
  the latter approach is less natural because it relies on an ordering on the induction
  variables.}
by defining $v_1\eq s_1,\ldots,v_n\eq s_n\prec v_1\eq t_1,\ldots,v_n\eq t_n$ 
iff $s_1\preceq t_1\wedge \ldots\wedge s_n\preceq t_n$ and $s_1\not=t_1\vee\ldots\vee s_n\not=t_n$. 
Constrained clauses are ordered lexicographically with priority on the constraint, i.e.\ $\clause C\alpha\prec\clause D\beta$ iff $\alpha\prec\beta$, or $\alpha=\beta$ and $C\prec D$. This ordering is not total on ground constrained clauses, e.g.\ the constrained clauses $\clause{\Box}{u\eq a,v\eq b}$ and $\clause{\Box}{u\eq b,v\eq a}$ are incomparable, but the ordering is strong enough to support our completeness results and an extension of the usual notion of redundancy to constrained clauses.

An ordering $\prec$ is \emph{well-founded} if there is no infinite chain $t_1\succ t_2\succ\ldots$, it has the \emph{subterm property} if $t[t']_p\succ t'$ for all $t,t'$ where $t[t']_p\not= t'$, and it is \emph{stable under substitutions} if $t\succ t'$ implies $t\sigma\succ t'\sigma$ for all $t,t'$ and all substitutions $\sigma$.
A \emph{reduction ordering} is a well-founded ordering that has the subterm property and is stable under substitutions.

\subsection*{Rewrite Systems}
A binary relation $\to$ on $\Terms(\Signature,X)$ is a \emph{rewrite relation} if $s\to t$ implies $u[s\sigma]\to u[t\sigma]$ for all terms $u\in\Terms(\Signature,X)$ and all substitutions $\sigma$.
By $\leftrightarrow$ we denote the symmetric closure of $\to$, and by $\stackrel{*}{\to}$ (and $\stackrel{*}{\leftrightarrow}$, respectively) we denote the reflexive and transitive closure of $\to$ (and $\leftrightarrow$).

A set $R$ of equations is called a \emph{rewrite system} with respect to a term ordering $\prec$ if $s\prec t$ or $t\prec s$ for each equation $s\eq t\in R$. Elements of $R$ are called \emph{rewrite rules}.
We also write $s\to t\in R$ instead of $s\eq t\in R$ if $s\succ t$.
By $\to_R$ we denote the smallest rewrite relation for which $s\to_Rt$ whenever $s\to t\in R$.
A term $s$ is \emph{reducible} by $R$ if there is a term $t$ such that $s\to_R t$, and \emph{irreducible} or \emph{in normal form} (with respect to $R$) otherwise. The same notions also apply to constraints instead of terms.

The rewrite system $R$ is \emph{ground} if all equations in $R$ are ground. It is \emph{terminating} if there is no infinite chain $t_0\to_R t_1\to_R\ldots$~, and it is \emph{confluent} if for all terms $t,t_1,t_2$ such that $t\to_R^* t_1$ and $t\to_R^* t_2$ there is a term $t_3$ such that $t_1\to_R^* t_3$ and $t_2\to_R^* t_3$.

\subsection*{Herbrand Interpretations}
\label{construction of I_N}
A \emph{Herbrand interpretation} over the signature $\Signature$ is a congruence on the ground terms $\Terms(\Signature)$, where the denotation of a term $t$ is the equivalence class of $t$. 

  We recall the construction of the special Herbrand interpretation $\perfect{N}$ derived from a set $N$ of (unconstrained) clauses \cite{BachmairGanzinger94}.
  Let $\prec$ be a well-founded reduction ordering that is total on ground terms. 
We use induction on the clause ordering $\prec$ to define ground rewrite systems $E_C$, $R_C$ and $I_C$ for ground clauses over $\Terms(\Signature)$ by $R_C=\bigcup_{C\succ C'}E_{C'}$, and $I_C=R_C^*$, i.e.~$I_C$ is the reflexive, transitive closure of $R_C$. Moreover, $E_C=\{s\to t\}$ if $C=\Gamma\to\Delta,s\eq t$ is a ground instance of a clause from $N$ such that
\begin{enumerate}
\item $s\eq t$ is a strictly maximal occurrence of an equation in $C$ and $s\succ t$,
\item $s$ is irreducible by $R_C$,
\item $\Gamma\subseteq I_C$, and
\item $\Delta\cap I_C=\emptyset$.
\end{enumerate}
In this case, we say that $C$ is \emph{productive} or that $C$ \emph{produces} $s\to t$. Otherwise $E_C=\emptyset$. Finally, we define a ground rewrite system $R_N=\bigcup_CE_C$ as the set of all produced rewrite rules and define the interpretation $\perfect{N}$ over the domain $\Terms(\Signature)$ as $\perfect{N}={\stackrel*\leftrightarrow_{R_N}}$.
The rewrite system $R_N$ is confluent and terminating. If $N$ is consistent and saturated with respect to a complete inference system then $\perfect{N}$ is a minimal model of $N$ with respect to set inclusion.

We will extend this construction of $\perfect{N}$ to constrained clauses in Section~\ref{sec:completeness}.

\subsection*{Constrained Clause Sets and Their Models}
If $V=\{v_1,\ldots,v_n\}$ and $N$ is a set of constrained clauses, then the semantics of $N$ is that there is a valuation of the existential variables, such that for all valuations of the universal variables, the constraint of each constrained clause in $N$ implies the respective clausal part:
An interpretation $\Model M$ \emph{models} $N$, written $\Model{M}\models N$, iff there is a map $\sigma$ from the set of existential variables to the universe of $\Model{M}$,\footnote{If $\Model{M}$ is a Herbrand interpretation over the signature $\Signature$, then $\sigma:V\to\Terms(\Signature)$ is a substitution mapping every existential variable to a ground term.} such that for each $(\clause{C}{\alpha})\in N$, the formula $\forall x_1,\ldots,x_m.\alpha\to C$ is valid in $\Model{M}$ under $\sigma$, where $\tenum x m$ are the universal variables of $\clause C\alpha$. In this case, $\Model{M}$ is called a \emph{model} of $N$. If $\Model{M}$ is also a Herbrand interpretation over the signature $\Signature$ of $N$, we call $\Model{M}$ a \emph{Herbrand model} of $N$.

For example, every Herbrand interpretation over the signature $\{0,s\}$ is a model of $\{\clause{\Box}{v\eq 0}\}$, because instantiating $v$ to $s(0)$ falsifies the constraint. On the other hand, the set $\{\clause{\Box}{v\eq 0},\ \clause{\Box}{v\eq s(x)}\}$ does not have any Herbrand models over $\{0,s\}$ because each instantiation of $v$ to a ground term over this signature validates one of the constraints, so that the corresponding constrained clause is falsified.

Note that the existential quantifiers range over the whole constrained clause set instead of each single constrained clause. The possibly most surprising effect of this is that two constrained clause sets may hold individually in a given interpretation while their union does not. As an example, note that the interpretation $\{P(s(0))\}$ models both constrained clause sets $\{\clause{P(x)}{v\eq x}\}$ (namely for $v\mapsto s(0)$) and $\{\clause{P(s(x))}{v\eq x}\}$ (namely for $v\mapsto 0$). However, the union $\{\clause{P(x)}{v\eq x},\ \clause{P(s(x))}{v\eq x}\}$ is not modeled by $\{P(s(0))\}$ because there is no instantiation of $v$ that is suitable for both constrained clauses.

Let $M$ and $N$ be two (constrained) clause sets.
We write $N\models M$ if each model of $N$ is also a model of $M$. We write $N\models_\Signature M$ if the same holds for each Herbrand model of $N$ over $\Signature$, and $N\models_\Ind M$ if $\perfect{N}\models M$.
A constrained clause set is \emph{satisfiable} if it has a model, and it is \emph{satisfiable over $\Signature$} if it has a Herbrand model over $\Signature$.

\subsection*{Inference Rules and Redundancy}
An \emph{inference rule} is a relation on constrained clauses. Its elements are called \emph{inferences} and are written as\footnote{Inference rules are sometimes marked by the letter $\mathcal{I}$ to differentiate them from reduction rules, marked by $\mathcal R$, where the premises are replaced by the conclusion. Since all rules appearing in this article are inference rules, we omit this marker.}
\[\inferit{\clause{C_1}{\alpha_1}\ \ldots\ 
  \clause{C_k}{\alpha_k}}{\clause{C}{\alpha}}{}\ .
\]
The constrained clauses $\clause{C_1}{\alpha_1},\ldots,\clause{C_k}{\alpha_k}$ are called the \emph{premises} and $\clause{C}{\alpha}$ the \emph{conclusion} of the inference.
  An \emph{inference system} is a set of inference rules.
  An inference rule is \emph{applicable} to a constrained clause set $N$ if the premises of the rule are contained in $N$. %

A ground constrained clause $\clause C\alpha$ is called \emph{redundant} with respect to a set $N$ of constrained clauses if there are ground instances $\clause{C_1}\alpha,\ldots,\clause{C_k}\alpha$ (with the common constraint $\alpha$) of constrained clauses in $N$ such that $C_i\prec C$ for all $i$ and $C_1,\ldots,C_k\models C$.\footnote{Note that $\models$ and $\models_\Signature$ agree on ground clauses over $\Signature$.}
A non-ground constrained clause is redundant if all its ground instances are redundant.
A ground inference with conclusion $\clause B\beta$ is called \emph{redundant} with respect to $N$ if either some premise is redundant or if there are ground instances $\clause{C_1}\beta,\ldots,\clause{C_k}\beta$ of constrained clauses in $N$ such that $C_1,\ldots,C_k\models B$ and $C_1,\ldots,C_n$ are smaller than the maximal premise of the ground inference.
A non-ground inference is redundant if all its ground instances are redundant.

A constrained clause set $N$ is \emph{saturated} (with respect to a given inference system) if each inference with premises in $N$ is redundant with respect to $N$.

\subsection*{Predicates}
Our notion of (constrained) clauses does not natively support predicative atoms. However, predicates can be included as follows:
We consider a many-sorted framework with two sorts $\textit{term}$ and $\textit{predicate}$, where the predicative sort is separated from the sort of all other terms.
The signature is extended by a new constant $\textit{true}$ of the predicative sort, and for each predicate $P$ by a function symbol $f_P$ of sort $\textit{term},\ldots,\textit{term}\to \textit{predicate}$. We then regard a predicative atom $P(t_1,\ldots,t_n)$ as an abbreviation for the equation $f_P(t_1,\ldots,t_n)\eq \textit{true}$.
As there are no variables of the predicative sort, substitutions do not introduce symbols of this sort and we never explicitly express the sorting, nor do we include predicative symbols when writing down signatures.

A given term ordering $\prec$ is extended to the new symbols such that $\textit{true}$ is minimal.

\section{First-Order Reasoning in Fixed Domains} \label{secforifx}

In this section, we will present a saturation procedure for sets of constrained clauses over a domain $\Terms(\Signature)$ and show how it is possible to decide whether a saturated constrained clause set possesses a Herbrand model over $\Signature$.
The calculus extends the superposition calculus of Bachmair and Ganzinger~\cite{BachmairGanzinger94}.

Before we come to the actual inference rules, let us review the semantics of constrained clauses by means of a simple example.
Consider the constrained clause set
\[\begin{array}{rl@{\ \|\ }rll}
    \{&\phantom{u\eq s(x),v\eq 0}& &\to G(s(x),0)&,\\
    &u\eq x, v\eq y   &G(x,y)   &\to&\}
  \end{array}\]
over the signature $\Signature_{\text{nat}}=\{s,0\}$.

This \TrTocl{set}{constrained clause set} corresponds to the formula
$\exists u,v. (\forall x. G(s(x),0)) \wedge \neg G(u,v)$.
In each Herbrand interpretation over $\Signature_{\text{nat}}$, this formula is equivalent to the formula
$\exists u,v. (\forall x. G(s(x),0))
 \wedge \neg G(u,v) \wedge (\forall x. u{\not=}s(x)\vee v{\not=}0)$, which corresponds to the following constrained clause set:
\[\begin{array}{rl@{\ \|\ }rll}
  \{&                 &         &\to G(s(x),0)&,\\
    &u\eq x, v\eq y   &G(x,y)   &\to &,\\
    &u\eq s(x),v\eq 0 & &\Box &\}
  \end{array}\]
Hence these two constrained clause sets are equivalent in every Herbrand interpretation over the signature $\Signature_{\text{nat}}$.

An aspect that catches the eye is that, although the clausal part of the last constrained clause is empty, this does not mean that the constrained clause set is unsatisfiable over $\Signature_{\text{nat}}$. The $\Box$ clause is constrained by $u\eq s(x)\wedge v\eq 0$, which means that, e.g., it is not satisfiable under the instantiation $u\mapsto s(0)$ and $v\mapsto 0$. In fact, the instantiated formula
$(\forall x. G(s(x),0)) \wedge \neg G(s(0),0) \wedge (\forall x. s(0){\not=}s(x)\vee 0{\not=}0)$
is unsatisfiable.
On the other hand, the clause set is satisfiable under the instantiation $u\mapsto 0$ and $v\mapsto s(0)$.

Derivations using our calculus will usually contain multiple, potentially infinitely many, constrained clauses with empty clausal parts. We explore in Theorem~\ref{Herbrand completeness} how the unsatisfiability of a saturated set of constrained clauses over $\Signature$ depends on a covering property of the constraints of constrained clauses with empty clausal part. In Theorem~\ref{covering decidable}, we prove that this property is decidable for finite constrained clause sets. Furthermore, we show how to saturate a given set of constrained clauses (Theorem~\ref{fair derivation makes saturation}).
Finally, we present in Section~\ref{sec:more general models} an extension of the calculus that allows to deduce a wider range of Herbrand models of $\Signature$-satisfiable constrained clause sets.

\subsection[The Superposition Calculus for Fixed Domains]{The Superposition \TrTocl{Calculus\\}{Calculus} for Fixed Domains}
\label{subseccalcnonskolem}

We consider the following inference rules, which are defined with respect to a reduction ordering $\prec$ on $\Terms(\Signature,X)$ that is total on ground terms.
Most of the rules are quite similar to the usual superposition rules~\cite{BachmairGanzinger94}, just generalized to constrained clauses. However, they require additional treatment of the constraints to avoid inferences like
\[\inferit{\clause{\to a\eq b}{u\eq f(x)}
           \quad \clause{a\eq c\to}{u\eq g(y)}}%
          {\clause{b\eq c\to}{u\eq f(x),u\eq g(y)}}%
          {}\]
the conclusion of which contains the existential variable $u$ more than once in its constraint and hence is not a constrained clause. 
In addition, there are two new rules that rewrite constraints.

To simplify the presentation below, we do not enrich the calculus by the use of a negative literal selection function as in~\cite{BachmairGanzinger94}, although this is also possible.
As usual, we consider the universal variables in different appearing constrained clauses to be renamed apart.
If $\alpha_1=v_1\eq s_1,\ldots,v_n\eq s_n$ and $\alpha_2=v_1\eq t_1,\ldots,v_n\eq t_n$ are two constraints, we write $\alpha_1\eq\alpha_2$ for the equations $s_1\eq t_1,\ldots,s_n\eq t_n$, and $\mgu(\alpha_1,\alpha_2)$ for the most general simultaneous unifier of $(s_1,t_1),\ldots,(s_n,t_n)$.
Note that $\alpha_1\eq\alpha_2$ does not contain any existential variables.

\begin{definition}
The \emph{superposition calculus for fixed domains \BaseCalc} consists of the following inference rules:
\begin{longitem}
\item \textbf{Equality Resolution:}
  \[\inferit{\clause{\Gamma,s\eq t\to \Delta}{\alpha}}%
            {(\clause{\Gamma\to\Delta}{\alpha})\sigma}%
            {}\]
  where (i) $\sigma=\mgu(s,t)$ and
  (ii) %
    $(s\eq t)\sigma$ is maximal in $(\Gamma,s\eq t\to\Delta)\sigma$.%
    \footnote{Note that we do not consider the constraint part for the maximality condition.}
\item \textbf{Equality Factoring:}
  \[\inferit{\clause{\Gamma\to \Delta,s\eq t,s'\eq t'}{\alpha}}%
            {(\clause{\Gamma,t\eq t'\to\Delta,s'\eq t'}{\alpha})\sigma}%
            {}\]
  where (i) $\sigma=\mgu(s,s')$,
  (ii) $(s\eq t)\sigma$ is maximal in $(\Gamma\to \Delta,s\eq t,s'\eq t')\sigma$, and
  (iii) $t\sigma\not\succeq s\sigma$
\item \textbf{Superposition, Right:}
  \[\inferit{\clause{\Gamma_1\to \Delta_1,l\eq r}{\alpha_1}
             \quad \clause{\Gamma_2\to \Delta_2,s[l']_p\eq t}{\alpha_2}}%
            {(\clause{\Gamma_1,\Gamma_2\to\Delta_1,\Delta_2,s[r]_p\eq t}%
                    {\alpha_1})\sigma_1\sigma_2}%
            {}\]
  where (i) $\sigma_1=\mgu(l,l')$,
    $\sigma_2=\mgu(\alpha_1\sigma_1,\alpha_2\sigma_1)$,
  (ii) %
    $(l\eq r)\sigma_1\sigma_2$ is strictly maximal in
    $(\Gamma_1\to \Delta_1,l\eq r)\sigma_1\sigma_2$
    and $(s\eq t)\sigma_1\sigma_2$ is strictly maximal in
    $(\Gamma_2\to \Delta_2,s\eq t)\sigma_1\sigma_2$,
  (iii) $r\sigma_1\sigma_2\not\succeq l\sigma_1\sigma_2$
    and $t\sigma_1\sigma_2\not\succeq s\sigma_1\sigma_2$, and
  (iv) $l'$ is not a variable.
\item \textbf{Superposition, Left:}
  \[\inferit{\clause{\Gamma_1\to \Delta_1,l\eq r}{\alpha_1}
             \quad \clause{\Gamma_2,s[l']_p\eq t\to \Delta_2}{\alpha_2}}%
            {(\clause{\Gamma_1,\Gamma_2,s[r]_p\eq t\to\Delta_1,\Delta_2}%
                     {\alpha_1})\sigma_1\sigma_2}%
            {}\]
  where (i) $\sigma_1=\mgu(l,l')$,
    $\sigma_2=\mgu(\alpha_1\sigma_1,\alpha_2\sigma_1)$,
  (ii) %
    $(l\eq r)\sigma_1\sigma_2$ is strictly maximal in
    $(\Gamma_1\to \Delta_1,l\eq r)\sigma_1\sigma_2$ and
    $(s\eq t)\sigma_1\sigma_2$ is maximal in
    $(\Gamma_2\to \Delta_2,s\eq t)\sigma_1\sigma_2$, 
  (iii) $r\sigma_1\sigma_2\not\succeq l\sigma_1\sigma_2$
    and $t\sigma_1\sigma_2\not\succeq s\sigma_1\sigma_2$, and
  (iv) $l'$ is not a variable.
\item \textbf{Constraint Superposition:}
  \[\inferit{\clause{\Gamma_1\to\Delta_1,l\eq r}{\alpha_1}
             \quad \clause{\Gamma_2\to\Delta_2}{\alpha_2[l']}}%
            {(\clause{\alpha_1\eq\alpha_2[r],\Gamma_1,\Gamma_2\to\Delta_1,\Delta_2}%
                     {\alpha_2[r]})\sigma}%
            {}\]
  where (i) $\sigma=\mgu(l,l')$,
  (ii) %
    $(l\eq r)\sigma$ is strictly maximal
    in $(\Gamma_1\to\Delta_1,l\eq r)\sigma$,
  (iii) $r\sigma\not\succeq l\sigma$, and
  (iv) $l'$ is not a variable.
\item \textbf{Equality Elimination:} %
  \[\inferit{\clause{\Gamma\to\Delta,l\eq r}{\alpha_1}
             \quad \clause\Box{\alpha_2[r']}}%
            {(\clause{\Gamma\to\Delta}%
                     {\alpha_1})\sigma_1\sigma_2}%
            {}\]
  where (i) $\sigma_1=\mgu(r,r')$,
    $\sigma_2=\mgu(\alpha_1\sigma_1,\alpha_2[l]\sigma_1)$,
  (ii) %
    $(l\eq r)\sigma_1\sigma_2$ is strictly maximal
    in $(\Gamma\to\Delta,l\eq r)\sigma_1\sigma_2$,
  (iii) $r\sigma_1\sigma_2\not\succeq l\sigma_1\sigma_2$, and
  (iv) $r'$ is not a variable.
\end{longitem}
\end{definition}

This inference system contains the standard universal superposition calculus as the special case when there are no existential variables at all present, i.e.~$V=\emptyset$ and all constraints are empty: The rules equality resolution, equality factoring, and superposition right and left reduce to their non-constrained counterparts and the constraint superposition and equality elimination rules become obsolete.

While the former rules are thus well-known, a few words may be in order to explain the idea behind constraint superposition and equality elimination. They have been introduced to make the calculus \emph{refutationally complete}, i.e.~to ensure that constrained clause sets that are saturated with respect to the inference system and that do not have a Herbrand model over the given signature always contain ``enough'' constrained empty clauses (cf.~Definition~\ref{def:coverage} and Theorem~\ref{Herbrand completeness}).

A notable feature of constraint superposition is how the information of both premise constraints is combined in the conclusion. Classically, the existential variables would be Skolemized and the constraint of a constrained clause would be regarded as part of its antecedent. In this setting, superpositions into the constraint part as considered here would not even require a specialized rule but occur naturally in the following form:
\[\inferit{\alpha_1,\Gamma_1\to\Delta_1,l\eq r
           \quad \alpha_2[l'],\Gamma_2\to\Delta_2}%
          {(\alpha_1,\alpha_2[r],\Gamma_1,\Gamma_2\to\Delta_1,\Delta_2)\sigma}%
          {}\]
Translated into the language of constrained clauses, the conclusion would, however, not be a well-formed constrained clause.
In most inference rules, we circumvent this problem by forcing a unification of the constraints of the premises, so that we can use an equivalent and admissible conclusion. For constraint superposition, this approach turns out to be too weak to prove Proposition~\ref{non-chosen instances are irrelevant}. Therefore, we instead replace $\alpha_1$ by $\alpha_1\eq\alpha_2[r]$ in this inference rule to regain an admissible constrained clause.

The resulting constraint superposition rule alone is not sufficient to obtain refutational completeness. Abstractly speaking, it only transfers information about the equality relation from the clausal part into the constraint part. For completeness, we also need a transfer the other way round. Once we find terms that cannot be solutions to the existentially quantified variables, we have to propagate this information through the respective equivalence classes in the clausal part. The result is the rule equality elimination, which deletes equations that are in conflict with the satisfiability of constrained empty clauses.

\smallskip
The rules constraint superposition and equality elimination are the main reason why \BaseCalc\ can manage theories that are not constructor-based, i.e.\ where the calculus cannot assume the irreducibility of certain terms.

\begin{example}
Constraint superposition and equality elimination allow to derive, e.g., $\clause{\Box}{u\eq b}$ from $\clause{\to a\eq b}{u\eq b}$ and $\clause{\Box}{u\eq a}$, although $u\eq a$ and $u\eq b$ are not
unifiable:

If $b\succ a$, then $\clause{\Box}{u\eq b}$ is derived by one step of equality elimination:
\[
\inferit{\clause{\to b\eq a}{u\eq b}
         \quad \clause{\Box}{u\eq a}}%
        {\clause\Box{u\eq b}}%
        {Equality Elimination}
\]
Otherwise, $\clause{\Box}{u\eq b}$ follows from a step of constraint superposition and the subsequent resolution of a trivial equality:
\[\inferit
    {\inferit{\clause{\to a\eq b}{u\eq b}
              \quad \clause{\Box}{u\eq a}}%
             {\clause{b\eq b\to}{u\eq b}}%
             {Constraint Superposition}
    }%
    {\clause{\Box}{u\eq b}}%
    {Equality Resolution}
\vspace{-1em}\]
\qed
\end{example}

When we work with predicative atoms in the examples, we will not make the translation into the purely equational calculus explicit. If, e.g., $P$ is a predicate symbol that is translated into the function symbol $f_{\!P}$, we write a derivation
\[\inferit{
  \inferit{\clause{\Gamma_1\!\to\! \Delta_1,f_{\!P}(s_1,\ldots,s_n)\eq \textit{true}}{\alpha_1\!}
           \ \ \clause{\Gamma_2,f_{\!P}(t_1,\ldots,t_n)\eq \textit{true}\to \Delta_2}{\alpha_2}}%
          {(\clause{\Gamma_1,\Gamma_2,\textit{true}\eq \textit{true}\to\Delta_1,\Delta_2}%
                   {\alpha_1})\sigma_1\sigma_2}%
          {\TOCLonly{Superposition}}}
          {(\clause{\Gamma_1,\Gamma_2\to\Delta_1,\Delta_2}%
                   {\alpha_1})\sigma_1\sigma_2}%
          {\TOCLonly{Equality Resolution}}
\]
consisting of a superposition into a predicative atom and the subsequent resolution of the atom $\textit{true}\eq\textit{true}$
in the following condensed form:
\[\inferit{\clause{\Gamma_1\to \Delta_1,P(s_1,\ldots,s_n)}{\alpha_1}
           \quad \clause{\Gamma_2,P(t_1,\ldots,t_n)\to \Delta_2}{\alpha_2}}%
          {(\clause{\Gamma_1,\Gamma_2\to\Delta_1,\Delta_2}%
                   {\alpha_1})\sigma_1\sigma_2}%
          {\TOCLonly{Superposition}}\]

\smallskip
\begin{example}
  \label{elevator example}
  For a simple example involving only superposition on predicative atoms,
  consider the clause set
  $N_E = \{\to G(a,p)$, $\to G(b,q)$, $G(a,x)\rightarrow C(a,x)$, $C(a,x)\rightarrow R(a,x)$,
  $G(b,x)\rightarrow R(b,x)\}$
  that describes the elevator example presented in the introduction, and additionally the two
  constrained clauses
  $\clause{R(y,x)\to}{u\eq x,v\eq y}$ and $\clause{\to G(y,x)}{u\eq x,v\eq y}$.
  These clauses state that there are a person and an elevator, such that the person can reach
  the ground floor but not the restaurant floor in this elevator.
  
  Assume a term ordering $\prec$ for which $G(y,x)\prec C(y,x)\prec R(y,x)$.
  With this ordering, the succedent is strictly maximal in each clause of $N_E$. Because
  superposition inferences always work on maximal atoms (condition (ii)), only two inferences
  between the given constrained clauses are possible:
  \[\inferit{C(a,x)\rightarrow R(a,x)\quad \clause{R(y,x)\to}{u\eq x,v\eq y}}%
            {\clause{C(a,x)\to}{u\eq x,v\eq a}}{}\]
  \[\inferit{G(b,x)\rightarrow R(b,x)\quad \clause{R(y,x)\to}{u\eq x,v\eq y}}%
            {\clause{G(b,x)\to}{u\eq x,v\eq b}}{}\]
  The first conclusion can now be superposed with the third clause of $N_E$:
  \[\inferit{G(a,x)\rightarrow C(a,x)\quad \clause{C(a,x)\to}{u\eq x,v\eq a}}%
            {\clause{G(a,x)\to}{u\eq x,v\eq a}}{}\]
  The last two conclusions can in turn be superposed with the constrained clause
  $\clause{\to G(y,x)}{u\eq x,v\eq y}$:
  \[\inferit{\clause{\to G(y,x)}{u\eq x,v\eq y}\quad \clause{G(a,x)\to}{u\eq x,v\eq a}}%
            {\clause{\Box}{u\eq x,v\eq a}}{}\]
  \[\inferit{\clause{\to G(y,x)}{u\eq x,v\eq y}\quad \clause{G(b,x)\to}{u\eq x,v\eq b}}%
            {\clause{\Box}{u\eq x,v\eq b}}{}\]
  Now the only remaining \BaseCalc\ inferences are those 
  between the constrained clauses \mbox{$\to G(a,p)$} and $\clause{G(a,x)\to}{u\eq x,v\eq a}$
  and between the constrained clauses $\to G(b,q)$ and $\clause{G(b,x)\to}{u\eq x,v\eq b}$.
  They both result in clauses that are
  redundant with respect to the last two conclusions. Hence the inferences themselves
  are redundant, which means that they do not give us any new information on the system,
  and we can ignore them.

  In order to present such a series of inferences in a more concise manner, we will write
  them down as follows, where all constrained clauses are indexed and premises to an
  inference are represented by their indices:
  \[\begin{array}{rl@{\ \|\ }rl}
       \text{clauses in $N_E$: }     1:&              &              &\to G(a,p)\\
                                     2:&              &              &\to G(b,q)\\
                                     3:&              &G(a,x)        &\to C(a,x)\\
                                     4:&              &C(a,x)        &\to R(a,x)\\
                                     5:&              &G(b,x)        &\to R(b,x)\\
    \text{additional clauses: }      6:&u\eq x,v\eq y &R(y,x)        &\to \\
                                     7:&u\eq x,v\eq y &              &\to G(y,x)\\
           \text{Superposition(4,6)}=8:&u\eq x,v\eq a &C(a,x)        &\to \\
           \text{Superposition(5,6)}=9:&u\eq x,v\eq b &G(b,x)        &\to \\
          \text{Superposition(3,7)}=10:&u\eq x,v\eq a &G(a,x)        &\to \\
          \text{Superposition(7,9)}=11:&u\eq x,v\eq b &              &\Box \\
         \text{Superposition(7,10)}=12:&u\eq x,v\eq a &              &\Box \\
  \vspace{-2em}\end{array}\]
  \qed
\end{example}

\subsection[Model Construction and Refutational Completeness]{Model \TrTocl{Construction\\}{Construction} and Refutational Completeness}
\label{sec:completeness}

By treating each constraint as a part of the antecedent, constrained clauses can be regarded as a special class of unconstrained clauses. Because of this, the construction of a Herbrand interpretation for a set of constrained clauses is strongly connected to the one for universal clause sets~\cite{BachmairGanzinger94}. The main difference is that we now have to account for existential variables before starting the construction.
To define a Herbrand interpretation $\perfect{N}$ of a set $N$ of constrained clauses, we proceed in two steps: First, we identify an instantiation of the existential variables that does not contradict any constrained clauses with empty clausal part, and then we construct the model of a set of unconstrained clause instances.

\begin{definition}[Coverage]
  \label{def:coverage}
  Given a set $N$ of constrained clauses, we denote the set of all constraints of
  constrained clauses in $N$ with empty clausal part by $A_N$,
  i.e.~$A_N=\set{\alpha}{(\clause\Box\alpha)\in N}$.
    We call $A_N$ \emph{covering} if every ground constraint over the given signature
  is an instance of a constraint in $A_N$.
  
  Furthermore, we distinguish one constraint $\alpha_N$: If $A_N$ is not covering, then let
  $\alpha_N$ be a minimal ground constraint with respect to $\prec$ such that $\alpha_N$ is
  not an instance of any constraint in $A_N$. Otherwise let $\alpha_N$ be arbitrary.
\end{definition}

\begin{definition}[Minimal Model of a Constrained Clause Set]
  Let $N$ be a set of constrained clauses with associated ground constraint $\alpha_N$.
  The Herbrand interpretation $\perfect{N}^{\alpha_N}$ is defined as the minimal model
  of the unconstrained clause set
  $\set{C\sigma}{(\clause C{\alpha})\in N \wedge \alpha\sigma=\alpha_N}$
  as described in Section~\ref{sec:preliminaries}.
  We usually do not mention $\alpha_N$ explicitly and write $\perfect{N}$
  for $\perfect{N}^{\alpha_N}$ if no ambiguities arise from this.
\end{definition}

Note that even if $A_N$ is not covering, $\alpha_N$ is usually not uniquely defined.
E.g.~for the constrained clause set $N=\{\clause\Box{u\eq 0,v\eq 0}\}$ over $\Signature=\{0,s\}$,
it holds that $A_N=\{(u\eq 0,v\eq 0)\}$ and both $\alpha_N^1=\{u\eq 0, v\eq s(0)\}$ and $\alpha_N^2=\{u\eq s(0), v\eq 0\}$ are valid choices. When necessary, this ambiguity can be avoided by using an ordering on the existential variables as a tie breaker.

While it is well known how the construction of $\perfect{N}$ works once $\alpha_N$ is given, it is not that obvious that it is decidable whether $A_N$ is covering and, if it is not, effectively compute $\alpha_N$.
This is, however, possible for finite $A_N$:

\begin{theorem}[Decidability of Finite Coverage]
  \label{covering decidable}
  Let $N$ be a set of constrained clauses such that $A_N$ is finite.
  It is decidable whether $A_N$ is covering, and $\alpha_N$ is computable if $A_N$ is not covering.
\end{theorem}
\begin{proof}
  Consider the formula
  \[ \Phi=\bigwedge_{(\clause\Box{v_1\eq t_1,\ldots,v_n\eq t_n})
     \in N}  
  {v_1\noteq t_1\vee\ldots\vee v_n\noteq t_n}\]
  and let $\{x_1,\ldots,x_m\}\subseteq X$ be the set of universal variables occurring in $\Phi$.
  The set $A_N$ is not covering if and only if the formula
  $\forall x_1,\ldots,x_m.\Phi$ is satisfiable in $\Terms(\Signature)$.
  Such so-called disunification problems have been studied among others by Comon and Lescanne
  \cite{ComonLescanne89}, who gave a terminating algorithm that eliminates the universal
  quantifiers from $\forall x_1,\ldots,x_m.\Phi$ and transforms the initial problem into a formula
  $\tenum[\vee] \Psi m$, $m\geq 0$, such that each $\Psi_j$ is of the shape
  \[ \Psi_j=\exists \vec u.v_1\eq s_1\wedge \ldots\wedge v_n\eq s_n
                              \wedge z_1\noteq s_1'\wedge\ldots\wedge z_k\noteq s_k'\ ,\]
  where $v_1,\ldots,v_n$ occur only once in each $\Psi_j$, the $z_i$ are variables and $z_i\not= s_i'$.
  This is done in such a way that (un-)satisfiability in $\Terms(\Signature)$ is preserved.
  The formula $\tenum[\vee] \Psi m$ is satisfiable in $\Terms(\Signature)$ if and only if
  the disjunction is not empty.
  All solutions can easily be read off from the formula.
\end{proof}

For saturated sets, the information contained in the constrained empty clauses is already sufficient to decide whether Herbrand models exist:
Specifically, we will now show that a saturated constrained clause set $N$ has a Herbrand model over $\Signature$ (namely $\perfect{N}$) if and only if $A_N$ is not covering. In this case, $\perfect{N}$ is a minimal model of $\set{C\sigma}{(\clause C{\alpha})\in N \wedge \alpha\sigma=\alpha_N}$, and we will also call it the \emph{minimal model of $N$ (with respect to $\alpha_N$)}.
  Observe, however, that for other choices of $\alpha_N$ there may be strictly smaller models of $N$ with respect to set inclusion: For $N=\{\clause{\to P(s(0))}{},\ \clause{\to P(x)}{u\eq x}\}$, we have $\alpha_N=u\eq 0$ and $\perfect{N}=\perfect{N}^{\alpha_N}=\{P(0),P(s(0))\}$, and $\perfect{N}$ strictly contains the model $\{P(s(0))\}$ of $N$ that corresponds to the constraint $u\eq s(0)$.

  Since $\perfect{N}$ is defined via a set of unconstrained clauses, it inherits all properties of minimal models of unconstrained clause sets. Above all, we will use the property that the rewrite system $R_N$ constructed in parallel with $\perfect{N}$ is confluent and terminating.

\begin{lemma}
  \label{irreducible choice}
  Let $N$ be saturated with respect to the inference system \BaseCalc. 
  If $A_N$ is not covering then $\alpha_N$ is irreducible by $R_N$.
\end{lemma}
\begin{proof}
  Assume contrary to the proposition that $A_N$ is not covering and $\alpha_N$ is reducible.
  Then there are a position $p$ and a rule $l\sigma\to r\sigma\in R_N$ produced by a ground
  instance $(\clause{\Lambda\to\Pi,l\eq r}{\beta})\sigma$ of a constrained clause
  $\clause{\Lambda\to\Pi,l\eq r}{\beta}\in N$,  such that 
  $l\sigma=\alpha_N|_p$.
  \par
  Because of the minimality of $\alpha_N$ and because $\alpha_N\succ \alpha_N[r\sigma]_p$,
  there must be a constrained clause $\clause \Box{\gamma}\in N$ and a substitution $\sigma'$
  such that $\gamma\sigma'\eq \alpha_N[r\sigma]_p$.
  Since by definition $\alpha_N$ is not an instance of $\gamma$,
  the position $p$ is a non-variable position of $\gamma$.
  Since furthermore $\beta\sigma=\alpha_N=\gamma\sigma'[l\sigma]_p$
  and $\sigma$ is a unifier of $\gamma|_p$ and $r$
  and $\gamma\sigma'|_p=r\sigma$,
  there is an equality elimination inference as follows:
  \[\inferit{\clause{\Lambda\to\Pi,l\eq r}{\beta}
      \quad \clause\Box{\gamma}}%
     {(\clause{\Lambda\to\Pi}%
              {\beta})\sigma_1\sigma_2}%
     {$\sigma_1=\mgu(\gamma|_p,r)$,
      $\sigma_2=\mgu(\beta\sigma_1,\gamma[l]_p\sigma_1)$}\]
  Because of the saturation of $N$, the ground instance
  \[\inferit{(\clause{\Lambda\to\Pi,l\eq r}{\beta})\sigma
      \quad (\clause\Box{\gamma})\sigma'}%
     {(\clause{\Lambda\to\Pi}%
              {\beta})\sigma}%
     {}\]
  of this derivation is redundant.
  The first premise cannot be redundant because it is productive; the second one cannot be
  redundant because there are no clauses that are smaller than $\Box$.  
  This means that the constrained clause $(\clause{\Lambda\to\Pi}{\beta})\sigma$
  follows from ground instances of constrained clauses in $N$ all of which are smaller
  than the maximal premise $(\clause{\Lambda\to\Pi,l\eq r}{\beta})\sigma$.
  But then the same ground instances imply
  $(\clause{\Lambda\to\Pi,l\eq r}{\beta})\sigma$, which means that this constrained clause
  cannot be productive. A contradiction.
\end{proof}

\begin{lemma}
  \label{non-chosen instances are irrelevant}
  Let $N$ be saturated with respect to \BaseCalc\ and let $A_N$ not be covering.
  If $\perfect{N}\not\models N$ and if $(\clause C\alpha)\sigma$ is a minimal ground instance
  of a constrained clause in $N$ such that
  $\perfect{N}\not\models (\clause C\alpha)\sigma$, then $\alpha\sigma=\alpha_N$.
\end{lemma}
\begin{proof}
  Let $C=\Gamma\to\Delta$.
  By definition of entailment, $\perfect{N}\not\models (\clause C\alpha)\sigma$
  implies that $\perfect{N}\models \alpha_N\eq \alpha\sigma$,
  or equivalently $\alpha_N\stackrel*\leftrightarrow_{R_N} \alpha\sigma$.
  We have already seen in Lemma~\ref{irreducible choice} that $\alpha_N$ is irreducible.
  Because of the confluence of $R_N$, either $\alpha\sigma=\alpha_N$
  or $\alpha\sigma$ must be reducible.
  
  Assume the latter, i.e.\ that $\alpha\sigma|_p=l\sigma'$ for a position $p$ and a rule
  $l\sigma'\to r\sigma'\in R_N$ that has been produced by the ground instance
  $(\clause{\Lambda\to\Pi,l\eq r}{\beta})\sigma'$
   of a constrained clause $\clause{\Lambda\to\Pi,l\eq r}{\beta}\in N$.
  If $p$ is a variable position in $\alpha$ or not a position in $\alpha$ at all,
  then the rule actually reduces $\sigma$,
  which contradicts the minimality of $(\clause C\alpha)\sigma$.
  Otherwise, there is a constraint superposition inference
  \[\inferit{\clause{\Lambda\to\Pi,l\eq r}{\beta}
      \quad \clause {\Gamma\to\Delta}\alpha}%
     {(\clause{\beta\eq\alpha[r]_p,\Lambda,\Gamma\to\Pi,\Delta}%
              {\alpha[r]_p})\tau}%
     {$\tau=\mgu(\alpha|_p,l)$\ .}\]
  Consider the ground instance
  $\clause D\delta:=(\clause{\beta\eq\alpha[r]_p,\Lambda,\Gamma\to\Pi,\Delta}{\alpha[r]_p})
   \sigma\sigma'$
  of the conclusion.
  This constrained clause is not modeled by $\perfect{N}$.
  On the other hand, that $N$ is saturated implies that the ground inference 
  \[\inferit{(\clause{\Lambda\to\Pi,l\eq r}{\beta})\sigma'
      \quad (\clause {\Gamma\to\Delta}\alpha)\sigma}%
     {(\clause{\beta\eq\alpha[r]_p,\Lambda,\Gamma\to\Pi,\Delta}%
              {\alpha[r]_p})\sigma\sigma'}%
     {}\]
  is redundant.
  The premises cannot be redundant, because $(\clause{\Lambda\to\Pi,l\eq r}{\beta})\sigma'$
  is productive and $(\clause C\alpha)\sigma$ is minimal, so the constrained clause $\clause D\delta$
  follows from ground instances of constrained clauses of $N$ all of which are smaller than
  $\clause D\delta$. Since moreover $\clause D\delta \prec (\clause C\alpha)\sigma$, all these
  ground instances hold in $\perfect{N}$, hence $\perfect{N}\models \clause D\delta$
  by minimality of $(\clause C\alpha)\sigma$.
  This is a contradiction to $\perfect{N}\not\models \clause D\delta$.
\end{proof}

\begin{proposition}
  \label{perfect model is a model}
  Let $N$ be a set of constrained clauses such that $N$ is saturated with respect to \BaseCalc\
  and $A_N$ is not covering.
  Then $\perfect{N}\models N$.
\end{proposition}
\begin{proof}
  Assume, contrary to the proposition, that $N$ is saturated, $A_N$ is not covering,
  and $\perfect{N}\not\models N$.
  Then there is a minimal ground instance $(\clause C\alpha)\sigma$ of a constrained clause
  $\clause C\alpha\in N$ that is not modeled by $\perfect{N}$.
  We will refute this minimality. We proceed by a case analysis of the position of the maximal
  literal in $C\sigma$. As usual, we assume that the appearing constrained clauses do not share any
  non-existential variables.
  \begin{longitem}
  \item $C=\Gamma,s\eq t\to\Delta$ and $s\sigma\eq t\sigma$ is maximal in $C\sigma$ with
    $s\sigma=t\sigma$.
    Then $s$ and $t$ are unifiable, and so
    there is an inference by equality resolution as follows:
    \[\inferit{\clause{\Gamma,s\eq t\to \Delta}{\alpha}}%
              {(\clause{\Gamma\to\Delta}{\alpha})\sigma_1}%
              {$\sigma_1=\mgu(s,t)$}\]
      Consider the ground instance  $(\clause{\Gamma\to\Delta}{\alpha})\sigma$ of the conclusion.
      From this constrained clause, a contradiction can be obtained as in the proof
      of Lemma~\ref{non-chosen instances are irrelevant}.
  \item $C=\Gamma,s\eq t\to\Delta$ and $s\sigma\eq t\sigma$ is maximal in
    $C\sigma$ with $s\sigma\succ t\sigma$.
    Since $\perfect{N}\not\models C\sigma$, we know that $s\sigma\eq t\sigma\in \perfect{N}$,
    and because $R_N$ only rewrites larger to smaller terms $s\sigma$ must be reducible by a rule
    $l\sigma'{\to} r\sigma'\in R_N$
    produced by a ground instance $(\clause{\Lambda\to\Pi,l\eq r}\beta)\sigma'$
    of a constrained clause $\clause{\Lambda\to\Pi,l\eq r}\beta\in N$.
    So $s\sigma|_p=l\sigma'$ for some position p in $s\sigma$.\newline
    Case 1: $p$ is a non-variable position in $s$.
      Since $\beta\sigma'=\alpha_N=\alpha\sigma$
      and $s\sigma|_p=l\sigma'$,
      there is an inference by left superposition as follows:
      \[\inferit{\clause{\Lambda\to\Pi,l\eq r}{\beta}
                 \quad \clause{\Gamma,s\eq t\to \Delta}{\alpha}}%
                {(\clause{\Lambda,\Gamma,s[r]_p\eq t\to\Pi,\Delta}%
                        {\alpha})\sigma_1\sigma_2}%
                {$\sigma_1:=\mgu(s|_p,l),\TOCLonly{\ }\sigma_2=\mgu(\beta\sigma_1,\alpha\sigma_1)$}\]
      As before, a contradiction can be derived from the existence of
      the ground instance
      $(\clause{\Lambda,\Gamma,s[r]_p\eq t\to\Pi,\Delta}\alpha)\sigma\sigma'$ of the conclusion.
      \newline
    Case 2: $p=p'p''$, where $s|_{p'}=x$ is a variable. Then $(x\sigma)|_{p''}=l\sigma$.
      If $\tau$ is the substitution that coincides with $\sigma$ except that
      $x\tau=x\sigma[r\sigma]_{p''}$, then $\perfect{N}\not\models C\tau$ and
      $(\clause C\alpha)\tau$ contradicts the minimality of $(\clause C\alpha)\sigma$.
  \item $C=\Gamma\to\Delta,s\eq t$ and $s\sigma\eq t\sigma$ is maximal in
    $C\sigma$ with $s\sigma=t\sigma$.
    This cannot happen because then $C\sigma$ would be a tautology.
  \item $C=\Gamma\to\Delta,s\eq t$ and $s\sigma\eq t\sigma$ is strictly maximal in
    $C\sigma$ with $s\sigma\succ t\sigma$.
    Since $\perfect{N}\not\models C\sigma$, we know that $\perfect{N}\models \Gamma\sigma$,
    $\perfect{N}\not\models \Delta\sigma$, and
    $\perfect{N}\not\models s\sigma\eq t\sigma$,
    and thus $C$ did not produce the rule $s\sigma\to t\sigma$.
    The only possible reason for this is that $s\sigma$ is reducible by a rule
    $l\sigma'{\to} r\sigma'\in R_N$
    produced by a ground instance $(\clause{\Lambda\to\Pi,l\eq r}\beta)\sigma'$
    of a constrained clause $\clause{\Lambda\to\Pi,l\eq r}\beta\in N$.
    So $s\sigma|_p=l\sigma'$ for some position p in $s\sigma$.\newline
    Case 1: $p$ is a non-variable position in $s$.
      Since $\beta\sigma'=\alpha_N=\alpha\sigma$
      and $s\sigma|_p=l\sigma'$,
      there is an inference by right superposition as follows:
      \[\inferit{\clause{\Lambda\to\Pi,l\eq r}{\beta}
                 \quad \clause{\Gamma\to \Delta,s\eq t}{\alpha}}%
                {(\clause{\Lambda,\Gamma\to\Pi,\Delta,s[r]_p\eq t}%
                        {\alpha})\sigma_1\sigma_2}%
                {$\sigma_1:=\mgu(s|_p,l),\TOCLonly{\ }\sigma_2=\mgu(\beta\sigma_1,\alpha\sigma_1)$}\]
      As before, a contradiction can be derived from the existence of
      the ground instance
      $(\clause{\Lambda,\Gamma\to\Pi,\Delta}\alpha,s[r]_p\eq t)\sigma\sigma'$ of the conclusion.
      \newline
    Case 2: $p=p'p''$, where $s|_{p'}=x$ is a variable. Then $(x\sigma)|_{p''}=l\sigma'$.
      If $\tau$ is the substitution that coincides with $\sigma$ except that
      $x\tau=x\sigma[r\sigma']_{p''}$, then $\perfect{N}\not\models C\tau$ and $C\tau$
      contradicts the minimality of $C\sigma$.
  \item $C=\Gamma\to\Delta,s\eq t$ and $s\sigma\eq t\sigma$ is maximal but not strictly maximal in
    $C\sigma$ with $s\sigma\succ t\sigma$. 
    Then $\Delta=\Delta',s'\eq t'$ such that $s'\sigma\eq t'\sigma$ is also maximal in $C\sigma$,
    i.e.\ without loss of generality $s\sigma=s'\sigma$ and $t\sigma=t'\sigma$.
    Then there is an inference by equality factoring as follows:
    \[\inferit{\clause{\Gamma\to \Delta',s\eq t,s'\eq t'}{\alpha}}%
              {(\clause{\Gamma,t\eq t'\to\Delta',s'\eq t'}{\alpha})\sigma_1}%
              {$\sigma_1=\mgu(s,s')$}\]
      In analogy to the previous cases, a contradiction can be derived from the existence of
      the ground instance
      $(\clause{\Gamma,t\eq t'\to\Delta',s'\eq t'}{\alpha})\sigma$ of the conclusion.
  \item $C\sigma$ does not contain any maximal literal at all, i.e.\ $C=\Box$.
    Since $\alpha\sigma=\alpha_N$ by Lemma~\ref{non-chosen instances are irrelevant} but
    $\perfect{N}\not\models\alpha\sigma\eq\alpha_N$ by definition of $\alpha_N$,
    this cannot happen.
  \end{longitem}
  Since we obtained a contradiction in each case, the initial assumption must be false, i.e.~the
  proposition holds.
\end{proof}

For the construction of $\perfect{N}$, we chose $\alpha_N$ to be minimal. For non-minimal $\alpha_N$, the proposition does not hold:

\begin{example}
  \label{bigger alphas are bad}
  If $N=\{\clause{\to a\eq b}{u\eq a},\ \clause{a\eq b\to}{u\eq b}\}$ and $a\succ b$,
  then no inference rule from \BaseCalc\ is applicable to $N$, so $N$ is saturated.
  However, $N$ implies $\clause\Box{u\eq a}$.
  So the Herbrand interpretation constructed with $\alpha_N'=\{u\eq a\}$ is not a model of $N$.\qed
\end{example}

On the other hand, whenever $N$ has any Herbrand model over $\Signature$ then $A_N$ is not covering:

\begin{proposition}
  \label{refutational completeness}
  Let $N$ be a set of cons\-trained clauses over $\Signature$ for which
  $A_N$ is covering.
  Then $N$ does not have any Herbrand model over $\Signature$.
\end{proposition}
\begin{proof}
  Let $\Model{M}$ be a Herbrand model of $N$ over $\Signature$.
  \\Then $\Model{M}\models\set{(\clause\Box\alpha)}{(\clause\Box\alpha)\in N}$, i.e.~there is
  a substitution $\sigma\In V\to \Terms(\Signature)$, such that for all $(\clause\Box\alpha)\in N$
  and all $\tau\In X\to \Terms(\Signature)$, $\Model M\models\alpha\sigma\tau$
  implies $\Model M\models\Box$. Since the latter is false, $\Model M\models \neg\alpha\sigma\tau$
  for all $\tau$, and so $\Model M\models \neg\alpha\sigma$.
  The same holds for the Herbrand model $\Model{M}_{\eq}$ over $\Signature$ where
  $\eq$ is interpreted as syntactic equality, i.e.~$\Model{M}_{\eq}\models \neg\alpha\sigma$.
  But then the constraint $\bigwedge_{v\in V}v\eq v\sigma$ is not an instance of the constraint
  of any constrained clause of the form $\clause\Box\alpha$,
  so $A_N$ is not covering.
\end{proof}

A constrained clause set $N$ for which $A_N$ is covering may nevertheless have both non-Herbrand models and Herbrand models over an extended signature:
If $\Signature=\{a\}$ and $N=\{\clause\Box{u\eq a}\}$ then $A_N$ is covering, but any standard first-order interpretation with a universe of at least two elements is a model of $N$.

\smallskip
Propositions~\ref{perfect model is a model} and~\ref{refutational completeness} constitute the following theorem:

\begin{theorem}[Refutational Completeness]
  \label{Herbrand completeness}
  Let $N$ be a set of constrained clauses over $\Signature$ that is saturated with respect to \BaseCalc.
  Then $N$ has a Herbrand model over $\Signature$ if and only if
  $A_N$ is not covering.
\end{theorem}

Moreover, the classical notions of (first-order) theorem proving derivations and fairness from \cite{BachmairGanzinger94} carry over to our setting.

\begin{definition}[Theorem Proving Derivations]
A (finite or countably infinite) \emph{$\models_\Signature$~theorem proving derivation} is a sequence $N_0,N_1,\ldots$ of constrained clause sets,
such that either
\begin{longitem}
\item (Deduction) $N_{i+1}=N_i\cup \{\clause C\alpha\}$
  and $N_i\models_\Signature N_{i+1}$, or
\item (Deletion) $N_{i+1}=N_i\setminus \{\clause C\alpha\}$ and $\clause C\alpha$ is redundant
       with respect to $N_i$.
\end{longitem}
If $N$ is a saturated constrained clause set for which $A_N$ is not covering, a \emph{$\models_\Ind$~theorem proving derivation for $N$} is a sequence $N_0,N_1,\ldots$ of constrained clause sets such that $N\subseteq N_0$ and either
\begin{longitem}
\item (Deduction) $N_{i+1}=N_i\cup \{\clause C\alpha\}$
  and $N\models_\Ind N_{i} \iff N\models_\Ind N_{i+1}$, or
\item (Deletion) $N_{i+1}=N_i\setminus \{\clause C\alpha\}$ and $\clause C\alpha$ is redundant
       with respect to $N_i$.
\end{longitem}
\end{definition}

Due to the semantics of constrained clauses and specifically the fact that all constrained clauses in a set are connected by common existential quantifiers, it does not suffice to require that $N_i\models_\Signature \clause C\alpha$ (or $N_i\models_\Ind \clause C\alpha$, respectively). E.g.\ for the signature $\Signature=\{a,b\}$ and $a\prec b$, the constrained clause $\clause C\alpha=\clause{\to x\eq b}{u\eq x}$ is modeled by every Herbrand interpretation over $\Signature$, but $\{\clause{\to x\eq a}{u\eq x}\}\not\models_\Ind \{\clause{\to x\eq a}{u\eq x}\}\cup\{\clause C\alpha\}$.

\medskip
Our calculus is sound, i.e.\ we may employ it for deductions in both types of theorem proving derivations:

\begin{lemma}
  \label{solutions do not change lemma}
  Let $\clause C\alpha$ be the conclusion of a \BaseCalc\ inference with premises in $N$.
  Then $N\tau\models N\tau\cup\{\alpha\tau\to C\tau\}$
  for each substitution $\tau\In V\to\Terms(\Signature)$.
\end{lemma}
\begin{proof}
  This proof relies on the soundness of paramodulation, the unordered correspondent
  to (unconstrained) superposition~\cite{NieuwenhuisRubio01handbook}.

  Let $\clause C\alpha$ be the conclusion of an inference from
  $\clause{C_1}{\alpha_1},\clause{C_2}{\alpha_2}\in N$.
  Then $\alpha\tau\to C\tau$ is (modulo (unconstrained) equality resolution) an instance of the
  conclusion of a paramodulation inference from $\alpha_1\tau\to C_1\tau$
  and $\alpha_2\tau\to C_2\tau$.
  Because of the soundness of the paramodulation rules, we have
  $N\tau\models N\tau\cup\{\alpha\tau\to C\tau\}$.
\end{proof}

\begin{proposition}[Soundness]
  \label{solutions do not change}%
  The calculus \BaseCalc\ is sound for $\models_\Signature$ and $\models_\Ind$
  theorem proving derivations:
  \begin{enumerate}
  \item Let $\clause C\alpha$ be the conclusion of a \BaseCalc\ inference with premises in $N$.
    Then $N\models_\Signature N\cup\{\clause C\alpha\}$.
  \item Let $N$ be saturated with respect to \BaseCalc, let $A_N$ not be covering,
    and let $\clause C\alpha$ be the conclusion of a \BaseCalc\ inference with premises in
    $N\cup N'$.
    Then $N\models_\Ind N'$ if and only if $N\models_\Ind N'\cup\{\clause C\alpha\}$.
  \end{enumerate}
\end{proposition}
\begin{proof}
  This follows directly from Lemma~\ref{solutions do not change lemma}.
\end{proof}

A $\models_\Signature$~or $\models_\Ind$~theorem proving derivation $N_0, N_1,\ldots$ is \emph{fair} if every inference with premises in the constrained clause set $N_\infty=\bigcup_j\bigcap_{k\geq j} N_k$ is redundant with respect to $\bigcup_j N_j$.
As usual, fairness can be ensured by systematically adding conclusions of non-redundant inferences, making these inferences redundant.

As it relies on redundancy and fairness rather than on a concrete inference system (as long as this system is sound), the proof of the next theorem is exactly as in the unconstrained case:

\begin{theorem}[Saturation]
  \label{fair derivation makes saturation}
  Let $N_0,N_1,N_2,\ldots$ be a fair $\models_\Signature$~theorem proving derivation.
  Then the set $N_\infty$ is saturated.
  Moreover, $N_0$ has a Herbrand model over $\Signature$ if and only if $N_\infty$ does.

  Let $N_0,N_1,\ldots$ be a fair $\models_\Ind$~theorem proving derivation for $N$.
  Then the set $N\cup N_\infty$ is saturated.
  Moreover, $N\models_\Ind N_0$ if and only if $N\models_\Ind N_\infty$.
\end{theorem}

\begin{example}
  Consider again the example of the elevator presented in the introduction.
  We will now prove that $\forall y,x.G(y,x)\rightarrow R(y,x)$ is valid in all
  Herbrand models of $N_E$ over $\{a,b\}$, i.e.~that
  $N_E\cup\{\neg\forall y,x.G(y,x)\rightarrow R(y,x)\}$ does not have any Herbrand
  models over $\{a,b\}$.
  Following the line of thought presented above, we transform the negated query into
  the constrained clause set
  \[ \{\ \clause{R(y,x) \to}{u\eq x,v\eq y},
    \quad\clause{\to G(y,x)}{u\eq x,v\eq y}\ \}\]
  and then saturate $N_E$ together with these clauses.
  This saturation is exactly what we did in Example~\ref{elevator example}.
  The derived constrained empty clauses are
  $\clause\Box{u\eq x,v\eq a}$ and $\clause\Box{u\eq x,v\eq b}$.
  Their constraints are covering for $\{a,b\}$, which means that the inital constrained
  clause set does not have any Herbrand models over $\{a,b\}$, i.e.~that
  $N_E \models_{\{a,b\}} \forall y,x.G(y,x)\rightarrow R(y,x)$.
  \qed
\end{example}

\subsection{Other Herbrand Models of Constrained Clause Sets}
\label{sec:more general models}

A so far open question in the definition of the minimal model $\perfect{N}$ is whether there is the alternative of choosing a non-minimal constraint $\alpha_N$. We have seen in Example~\ref{bigger alphas are bad} that this is in general not possible for sets $N$ that are saturated with respect to our present calculus, but we have also seen after Theorem~\ref{covering decidable} that models corresponding to non-minimal constraints may well be of interest.
Such a situation will occur again in Example~\ref{ex:anotherpartdef}, where knowledge about all models allows to find a complete set of counterexamples to a query.

To include also Herbrand models arising from non-minimal constraints, we now change our inference system. 
The trade-off is that we introduce a new and prolific inference rule that may introduce constrained clauses that are larger than the premises. This makes even the saturation of simple constrained clause sets non-terminating. E.g.\ a derivation starting from $\{\clause{\to f(a)\eq a}{},\ \clause{\to P(a)}{u\eq a}\}$ will successively produce \TOCLonly{the }increasingly large constrained clauses $\clause{\to P(a)}{u\eq f(a)}$, $\clause{\to P(a)}{u\eq f(f(a))}$ and so on.

The following two changes affect only this section.

\begin{definition}
  The calculus \emph{\GeneralCalc} arises from \BaseCalc\ by replacing the
  equality elimination inference rule by the following more general rule:
    \[\inferit{\clause{\Gamma\to\Delta,l\eq r}{\alpha_1}
               \quad \clause{\Gamma_2\to\Delta_2}{\alpha_2[r']}}%
              {(\clause{\alpha_1\eq\alpha_2[l],\Gamma_1,\Gamma_2\to\Delta_1,\Delta_2}%
                       {\alpha_2[l]})\sigma}%
              {}\]
    where (i) $\sigma=\mgu(r,r')$,
    (ii) $l\sigma\eq r\sigma$ is strictly maximal in
      $(\Gamma\to\Delta,l\eq r)\sigma$,
    (iii) $r\sigma\not\succeq l\sigma$, and
    (iv) $r'$ is not a variable.
\end{definition}

Note that in a purely predicative setting, i.e.~when all equations outside constraints are of the form $t\eq \textit{true}$, the separation of base sort and predicative sort prevents the application of both the original and the new equality elimination rule.
So the calculi \BaseCalc\ and \GeneralCalc\ coincide in this case.

\begin{definition}
  Let $N$ be a set of constrained clauses.
  If $A_N$ is not covering, then let $\alpha_N$ be \emph{any} ground constraint that is
  not an instance of any constraint in $A_N$ (note that $\alpha_N$ does not have to be minimal).
  Otherwise let $\alpha_N$ be arbitrary.

  The Herbrand interpretation $\perfect{N}^{\alpha_N}$ is still defined as the minimal model
  of the unconstrained clause set
  $\set{C\sigma}{(\clause C{\alpha})\in N \wedge \alpha\sigma=\alpha_N}$.
\end{definition}

Since the proof of Lemma~\ref{irreducible choice} depends strongly on the minimality of $\alpha_N$, we have to change our proof strategy and cannot rely on previous results.

\begin{lemma}
  \label{there are good counterexamples}
  Let $N$ be saturated with respect to \GeneralCalc. Assume that $A_N$ is not covering and fix
  some $\alpha_N$. If $\perfect{N}\not\models N$, then there is a ground instance
  $(\clause C\alpha)\sigma$ of a constrained clause in $N$ such that
  $\perfect{N}\not\models (\clause C\alpha)\sigma$ and $\alpha\sigma=\alpha_N$.
\end{lemma}
\begin{proof}
  Let $(\clause C\alpha)\sigma$ be the minimal ground instance of a constrained clause in $N$
  such that $\perfect{N}\not\models (\clause C\alpha)\sigma$.
  We first show that we can restrict ourselves to the case where $\alpha_N$ rewrites to
  $\alpha\sigma$ using $R_N$ and then solve this case.
  
  \vspace{0.5em}
  $\perfect{N}\not\models (\clause C\alpha)\sigma$ implies
  $\perfect{N}\models \alpha\sigma\eq \alpha_N$,
  thus by confluence of $R_N$
  \[\alpha\sigma\stackrel*\rightarrow_{R_N}\alpha_0\phantom{t}_{R_N}
    \stackrel*\leftarrow \alpha_N\ ,\]
  where $\alpha_0$ is the normal form of $\alpha_N$ under $R_N$.
  We show that $\alpha\sigma=\alpha_0$.
    
  If $\alpha\sigma\not=\alpha_0$, then there is a rule $l\sigma'\to r\sigma'\in R_N$
  that was produced by the ground instance $(\clause {\Lambda\to\Pi,l\eq r}\beta)\sigma'$
  of a constrained clause $\clause {\Lambda\to\Pi,l\eq r}\beta\in N$ such that
  $\alpha\sigma[l\sigma']_p\rightarrow_{R_N}\alpha\sigma[r\sigma']_p$.
    
  If $p$ is a variable position in $\alpha$ or not a position in $\alpha$ at all,
  then the rule actually reduces $\sigma$,
  which contradicts the minimality of $(\clause C\alpha)\sigma$.

  So $p$ must be a non-variable position of $\alpha$.
  Let $C=\Gamma\to\Delta$. Then there is a constraint superposition inference as follows:
  \[\inferit{\clause{\Lambda\to\Pi,l\eq r}{\beta}
      \quad \clause{\Gamma\to\Delta}\alpha}%
     {(\clause{\beta\eq\alpha[r]_p,\Lambda,\Gamma\to\Pi,\Delta}%
              {\alpha[r]_p})\tau}%
     {$\tau=\mgu(\alpha|_p,l)$}\]
  The ground instance
  $\clause D\delta:=(\clause{\beta\eq\alpha[r]_p,\Lambda,\Gamma\to\Pi,\Delta}
      {\alpha[r]_p})\sigma\sigma'$
  of the conclusion is not modeled by $\perfect{N}$.
  On the other hand, because $N$ is saturated, the ground instance
  \[\inferit{(\clause{\Lambda\to\Pi,l\eq r}{\beta})\sigma'
      \quad (\clause{\Gamma\to\Delta}\alpha)\sigma}%
     {(\clause{\beta\eq\alpha[r]_p,\Lambda,\Gamma\to\Pi,\Delta}%
              {\alpha[r]_p})\sigma\sigma'}%
     {}\]
  of the above inference is redundant.
  The first premise cannot be redundant because it is productive; the second one cannot be
  redundant because of the minimality of $(\clause{\Gamma\to\Delta}\alpha)\sigma$.
  This means that the conclusion
  follows from ground instances of constrained clauses in $N$ all of which are smaller
  than the maximal premise $(\clause{\Gamma\to\Delta}\alpha)\sigma$.
  All these
  ground instances are modeled by $\perfect{N}$, and so $\perfect{N}\models \clause D\delta$.

  \vspace{1ex}
  So whenever $\perfect{N}\not\models N$, there is a ground instance
  $(\clause C\alpha)\sigma$ of a constrained clause in $N$ such that
  $\perfect{N}\not\models (\clause C\alpha)\sigma$ and
  $\alpha\sigma=\alpha_0$. In particular $\alpha_N\stackrel*\rightarrow_{R_N}\alpha\sigma$.

  \vspace{1ex}
  Let $n\in\Nat$ be the minimal number for which there is a ground instance
  $(\clause C\alpha)\sigma$ of a constrained clause
  $\clause C\alpha=\clause {\Gamma\to\Delta}\alpha$ in $N$ such that
  $\perfect{N}\not\models (\clause C\alpha)\sigma$ and $\alpha_N$ rewrites to $\alpha\sigma$
  via $R_N$ in $n$ steps, written $\alpha_N\to_{R_N}^n\alpha\sigma$. We have to show that $n=0$.
    
  Assume $n>0$. Then
  the last step of the derivation $\alpha_N\to_{R_N}^n\alpha\sigma$ is of the form
  $\alpha\sigma[l\sigma']_p\rightarrow_{R_N}\alpha\sigma[r\sigma']_p=\alpha\sigma$,
  where the rule $l\sigma'\to r\sigma'\in R_N$
  has been produced by a constrained clause
  $\clause {\Lambda\to\Pi,l\eq r}\beta\in N$ with $\beta\sigma'=\alpha_N$.
    
  If $p$ is a variable position in $\alpha$ or not a position in $\alpha$ at all,
  we write $p=p'p''$ such that $\alpha|_{p'}=x$ is a variable.
  Let $\tau$ be the substitution that coincides with $\sigma$ except that
  $x\tau=x\sigma[l\sigma']_{p''}$. Then $\perfect{N}\not\models (\clause C\alpha)\tau$ and
  $\alpha_N\to_{R_N}^{n-1}\alpha\tau$ contradicts the minimality of $n$.

  Otherwise there is an equality elimination inference as follows:
  \[\inferit{\clause{\Lambda\to\Pi,l\eq r}{\beta}
      \quad \clause{\Gamma\to\Delta}\alpha}%
     {(\clause{\beta\eq\alpha[l]_p,\Lambda,\Gamma\to\Pi,\Delta}%
              {\alpha[l]_p})\tau}%
     {$\tau=\mgu(\alpha|_p,r)$}\]
  The ground instance
  $\clause D\delta:=(\clause{\beta\eq\alpha[l]_p,\Lambda,\Gamma\to\Pi,\Delta}
                            {\alpha[l]_p})\sigma\sigma'$
  of the conclusion is not modeled by $\perfect{N}$.
  In particular, $\perfect{N}\models \delta$ and $\perfect{N}\not\models D$.
    
  Since the inference, and hence also the constrained clause $\clause D\delta$ is redundant,
  there are constrained clauses $\clause {D_1}{\delta_1},\ldots,\clause {D_m}{\delta_m}\in N$
  together with substitutions $\sigma_1,\ldots,\sigma_m$,
  such that $\delta=\delta_i\sigma_i$ for all $i$
  and $D_1\sigma_1,\ldots,D_m\sigma_m\models D$. This implies that
  $\perfect{N}\not\models(\clause{D_i}{\delta_i})\sigma_i$ for at least one of the
  constrained clause instances $(\clause{D_i}{\delta_i})\sigma_i$.
  Since $\alpha_N\to_{R_N}^{n-1}\delta_i\sigma_i=\delta=\alpha\sigma[l\sigma']$,
  this contradicts the minimality of $n$.
\end{proof}

With this preparatory work done, we can reprove Proposition~\ref{refutational completeness} and Theorem~\ref{Herbrand completeness} in this new setting:

\begin{proposition}
  \label{general perfect model is a model}
  Let $N$ be a set of constrained clauses that is saturated with respect to \GeneralCalc.
  Then $\perfect{N}^{\alpha_N}\models N$ for any ground constraint $\alpha_N$
  that is not covered by $A_N$.
\end{proposition}
\begin{proof}
  The proof is almost identical to the proof of Proposition~\ref{perfect model is a model}.
  The only difference is that, instead of reasoning about
  the minimal ground instance $(\clause C\alpha)\sigma$
  of a constrained clause $\clause C\alpha\in N$ that is not modeled by $\perfect{N}$,  
  we consider the minimal such instance that
  additionally satisfies $\alpha\sigma=\alpha_N$.
  Lemma~\ref{there are good counterexamples} states that this is sufficient.
\end{proof}

\begin{theorem}[Refutational Completeness]
  \label{generalized Herbrand completeness}
  Let $N$ be a set of constrained clauses over $\Signature$ that is saturated with respect to
  \GeneralCalc.
  Then $N$ has a Herbrand model over $\Signature$ if and only if
  $A_N$ is not covering.
\end{theorem}

\section{Fixed Domain and Minimal Model Validity\\
         of Constrained Clauses}
\label{sec: fixdom and minmod validity}

Given a constrained or unconstrained clause set $N$, we are often not only interested in the (un)satisfiability of $N$ (with or without respect to a fixed domain), but also in properties of Herbrand models of a $N$ over $\Signature$, especially of $\perfect{N}$.
  These are not always disjoint problems: We will show in Proposition~\ref{N*=Nind for exists Gamma} that, for some $N$ and queries of the form $\exists \vec x.\tenum[\wedge] A n$, first-order validity and validity in $\perfect{N}$ coincide, so that we can explore the latter with first-order techniques.

The result can be extended further: We will use our superposition calculus \BaseCalc\ to
demonstrate classes of constrained clause sets $N$ and $H$ for which $N\models_\Signature H$ and $N\models_\Ind H$ coincide (Proposition~\ref{models_ind = models_sig for positive universal clauses}).
  Finally, we will look at ways to improve the termination of our approach for proving properties of $\perfect{N}$ (Theorem~\ref{induction rule is sound}).

In this context, it is important to carefully observe the semantics of, e.g., the expression $N\models_\Ind H$ when $N$ is constrained.
Consider for example the signature $\Signature = \{a,b\}$ with $a\succ b$, $N_P=\{\clause{\to P(x)}{u\eq x}\}$ and $H_P=\{\clause{P(x)\to}{u\eq x}\}$. Then $N_P\cup H_P$ is unsatisfiable, but nevertheless $H_P$ is valid in the model $\perfect{N_P}=\{P(b)\}$, i.e.\ $N_P\models_\Ind H_P$.
These difficulties vanish when the existential variables in $N_P$ and $H_P$ are renamed apart.

\subsection{Relations between $\models$, $\models_\Signature$, and $\models_\Ind$}

Even with standard first-order superposition, we can prove that first-order validity and validity in $\perfect{N}$ coincide for some $N$ and properties $\Gamma$:

\begin{proposition}
  \label{N*=Nind for exists Gamma}
  If $N$ is a saturated set of unconstrained Horn clauses
  and $\Gamma$ is a conjunction of positive literals
  with existential closure $\exists \vec x.\Gamma$, then
  \begin{align*}
    N \models_\Ind \exists \vec x. \Gamma \iff N \models \exists \vec x. \Gamma\ .
  \end{align*}
\end{proposition}
\begin{proof}
  $N\models\exists \vec x.\Gamma$ holds
  if and only if the set $N\cup\{\forall \vec x.\neg\Gamma\}$ is unsatisfiable.
  $N$ is Horn, so during saturation of $N\cup\{\neg\Gamma\}$, where inferences between clauses
  in $N$ need not be performed,
  only purely negative, hence non-productive, clauses can appear.
  That means that the Herbrand interpretation $\perfect{N'}$ is the same for every
  clause set $N'$ in the derivation.
  So $N\cup\{\neg \Gamma\}$ is unsatisfiable if and only if
  $N \not\models_\Ind \forall \vec x.\neg \Gamma$, which is in turn equivalent to
  $N \models_\Ind \exists \vec x. \Gamma$.
\end{proof}

If $N$ and $\Gamma$ additionally belong to the Horn fragment of a first-order logic (clause) class decidable by (unconstrained) superposition, such as for example the monadic class with equality~\cite{BachmairGanzingerEtAl93a} or the guarded fragment with equality~\cite{GanzingerNeville99}, it is thus decidable whether $N \models_\Ind \exists \vec x. \Gamma$.

Given our superposition calculus for fixed domains, we can show that a result similar to Proposition~\ref{N*=Nind for exists Gamma} holds for universally quantified queries.

\begin{proposition}
  \label{models_ind = models_sig for positive universal clauses}
  If $N$ is a saturated set of Horn clauses
  and $\Gamma$ is a conjunction of positive literals with universal closure
  $\forall \vec v.\Gamma$, then
  \begin{align*}
    N\models_\Ind \forall \vec v.\Gamma
    \iff N\models_\Signature \forall \vec v. \Gamma\ .
  \end{align*}
\end{proposition}
\begin{proof}
  $N\models_\Signature\forall \vec v.\Gamma$ holds
  if and only if $N\cup\{\exists \vec v.\neg\Gamma\}$ does not have a Herbrand model
  over $\Signature$.
  \par
  If $N\cup\{\exists \vec v.\neg\Gamma\}$ does not have a Herbrand model over $\Signature$,
  then obviously $N \not\models_\Ind \exists \vec v. \neg\Gamma$.

  Otherwise, consider the constrained clause $\clause{\Delta\to}{\alpha}$ corresponding to
  the formula $\exists \vec v. \neg\Gamma$ and
  assume without loss of generality that the existential variables in $N$ and $\alpha$
  are renamed apart.
  The minimal models of the two sets $N$ and $N\cup\{\clause{\Delta\to}{\alpha}\}$
  are identical, since during the saturation of $N\cup\{\clause{\Delta\to}{\alpha}\}$
  inferences between clauses in $N$ need not be performed and so
  only purely negative, hence non-productive, constrained clauses can be derived.
  This in turn just means that $N\models_\Ind\exists \vec v.\neg\Gamma$.
\end{proof}

These propositions can also be proved using agruments from model theory. The shown proofs using superposition or \BaseCalc, respectively, notably the argument about the lack of new productive clauses, illustrate recurring crucial concepts of super\-position-based inductive theorem proving. We will see in Example~\ref{ex:anotherpartdef} that other superposition-based algorithms often fail because they cannot obviate the derivation of productive clauses.

\begin{example}
  \label{ex:diverges}
  We consider the partial definition of the usual ordering on the naturals given
  by $N_G=\{{\to G(s(0),0)},\ {G(x,y)\to G(s(x),s(y))}\}$, as shown in the introduction.
  We want to use Proposition~\ref{models_ind = models_sig for positive universal clauses}
  to check whether or not $N_G\models_{\Signature_{\text{nat}}}\forall x.G(s(x),x)$.
  The first steps of a possible derivation are as follows:
  \[\begin{array}{rl@{\ \|\ }rl}
    \text{clauses in $N$: }\hfill    1:&         &         &\to G(s(0),0)\\
                                     2:&         &G(x,y)   &\to G(s(x),s(y))\\
    \text{negated conjecture: }\hfill3:&u\eq x   &G(s(x),x)&\to \\
           \text{Superposition(1,3)}=4:&u\eq 0   &         &\Box\\
           \text{Superposition(2,3)}=5:&u\eq s(y)&G(s(y),y)&\to \\
           \text{Superposition(1,5)}=6:&u\eq s(0)&         &\Box\\
           \text{Superposition(2,5)}=7:&u\eq s(s(z))&G(s(z),z)&\to
  \end{array}\]
  In the sequel, we repeatedly superpose the constrained clauses~1 and~2 into (descendants of)
  the constrained clause 5. This way, we successively derive all constrained clauses of the forms
  $\clause{G(s(x),x)\to}{u\eq s^n(x)}$
  and $\clause{\Box}{u\eq s^n(0)}$,
  where $s^n(0)$ denotes the $n$-fold application $s(\ldots s(s(0))\ldots)$ of $s$ to $0$,
  and analogously for $s^n(x)$.
  Since the constraints of the derived constrained $\Box$ clauses are covering in the limit,
  we know that $N_G\models_{\Signature_{\text{nat}}}\forall x.G(s(x),x)$. \qed
\end{example}

Using Proposition~\ref{models_ind = models_sig for positive universal clauses}, we can employ the calculus \BaseCalc\ for fixed domain reasoning to also decide properties of minimal models. This is even possible in cases for which neither the approach of Ganzinger and Stuber~\cite{GanzingerStuber92} nor the one of Comon and Nieuwenhuis~\cite{ComonNieuwenhuis00} works.

\begin{example}
  \label{ex:anotherpartdef}
  Consider \TrTocl{a}{yet another} partial definition of the usual ordering on the naturals given
  by the saturated set $N_G'=\{{\to G(s(x),0)},\ {G(x,s(y))\to G(x,0)}\}$
  over the signature $\Signature_{\text{nat}}=\{0,s\}$.
  We want to prove %
  $N_G'\not\models_{\Ind}\forall x,y.G(x,y)$.
  \begin{longitem}
\item We start with the constrained clause $\clause{G(x,y)\to}{u\eq x,v\eq y}$ and do the following
  one step derivation:
  \[\begin{array}{rl@{\ \|\ }rl}
         \text{clauses in $N$:}\hfill1:&                 &         &\to G(s(x),0)\\
                                     2:&                 &G(x,s(y))&\to G(x,0)\\
    \text{negated conjecture: }\hfill3:&u\eq x, v\eq y   &G(x,y)   &\to \\
           \text{Superposition(1,3)}=4:&u\eq s(x),v\eq 0 & &\Box
  \end{array}\]
  All further inferences are redundant (even for the extended calculus \GeneralCalc\ from
  Section~\ref{sec:more general models}), thus the counter examples to the query are exactly
  those for which no constrained empty clause was derived, i.e.\ instantiations of $u$ and $v$
  which are not an instance of $\{u\mapsto s(x), v\mapsto 0\}$. Hence, these counter examples take
  on exactly the form $\{u\mapsto 0,v\mapsto t_2\}$ or $\{u\mapsto t_1,v\mapsto s(t_2)\}$ for any
  $t_1,t_2\in\Terms(\Signature_{\text{nat}})$.
  Thus we know that $N_G'\not\models_{\Signature_{\text{nat}}}\forall x,y.G(x,y)$, and since
  the query is positive, we also know that $N_G'\not\models_{\Ind}\forall x,y.G(x,y)$.
\item In comparison, the algorithm by Ganzinger and Stuber starts a derivation with the clause
  $\to G(x,y)$,
  derives in one step the potentially productive clause $\to G(x,0)$ and terminates
  with the answer ``don't know''.
  \par
  Ganzinger and Stuber also developed an extended approach that uses a predicate $\gnd$ defined by
  $\{\to\gnd(0),\ \gnd(x)\to\gnd(s(x))\}$. In this context, they guard each free variable $x$
  in a clause of $N$ and the conjecture by a literal $\gnd(x)$ in the antecedent.
  These literals mimic the effect of restricting the instantiation of variables to
  ground terms over $\Signature_{\text{nat}}$.
  The derivation then starts with the following clause set:
  \[\begin{array}{lrl}
    \text{clauses defining $\gnd$:\hspace{-5ex}}
      &&\to\gnd(0)\\
      &\gnd(x)&\to\gnd(s(x))\\
    \text{modified $N$:}
      &\gnd(x)&\to G(s(x),0)\\
      &\gnd(x),\gnd(y),G(x,s(y))&\to G(x,0)\\
    \text{conjecture:}
      &\gnd(x),\gnd(y)&\to G(x,y)
    \end{array}
  \]
  Whenever the conjecture or a derived clause contains negative $\gnd$ literals, one of these
  is selected, e.g.~always the leftmost one.
  This allows a series of superposition inferences with the clause $\gnd(x)\to\gnd(s(x))$,
  deriving the following infinite series of clauses:
  \[\begin{array}{@{\hspace{13.7em}}r@{\ }l}
      \gnd(x),\gnd(y)&\to G(x,y)\\
      \gnd(x_1),\gnd(y)&\to G(s(x_1),y)\\
      \gnd(x_2),\gnd(y)&\to G(s(s(x_2)),y)\\
      &\!\ldots
    \end{array}
  \]
  The extended algorithm diverges without producing an answer to the query.
\item The approach by Comon and Nieuwenhuis fails as well. Before starting the actual derivation,
  a so-called $I$-axiomatization of the negation of $G$ has to be computed. This involves a
  quantifier elimination procedure as in \cite{ComonLescanne89},
  that fails since the head of the clause $G(x,s(y))\to G(x,0)$ does not contain all variables
  of the clause:
  $G$ is defined in the minimal model $\perfect{N_G'}$ by
  $G(x,y) \iff (y=0\wedge \exists z.x=s(z))\vee(y=0\wedge \exists z.G(x,s(z)))$,
  so its negation is defined by
  $\neg G(x,y)\iff(y\not=0\vee \forall z.x\not=s(z))\wedge(y\not=0\vee \forall z.\neg G(x,s(z)))$.
  Quantifier elimination simplifies this to
  $\neg G(x,y)\iff (y\not=0\vee x=0)\wedge(y\not=0\vee \forall z.\neg G(x,s(z)))$
  but cannot get rid of the remaining universal quantifier:
  \[\begin{array}{rl@{\ }r}
    G(x,y) \iff& (y=0\wedge \exists z.x=s(z))\\
        &\phantom{}\vee(y=0\wedge \exists z.G(x,s(z)))\\
    \TRonly{\!\!\!}\neg G(x,y)
    \iff&(y\not=0\vee \forall z.x\not=s(z))\\
        &\phantom{}\wedge(y\not=0\vee \forall z.\neg G(x,s(z)))\\
    \iff&(y\not=0\vee (\forall z.x\not=s(z)\wedge x=0)
          &\text{by }E(x)\\
        &\quad\phantom{}\vee(\exists w.\forall z.x\not=s(z)\wedge x=s(w)))\\
        &\phantom{}\wedge(y\not=0\vee \forall z.\neg G(x,s(z)))\\
    \iff&(y\not=0\vee x=0
          &\!\!\!\text{by }R_1,C_2,\textit{UE}_1,{\to_{\mathcal{E}}}\\
        &\quad\phantom{}\vee(\exists w.\forall z.x\not=s(z)\wedge x=s(w)))\\
        &\phantom{}\wedge(y\not=0\vee \forall z.\neg G(x,s(z)))\\
    \iff&(y\not=0\vee x=0)
          &\text{by }R_1,D_2,\textit{UE}_2,\\
        &\phantom{}\wedge(y\not=0\vee \forall z.\neg G(x,s(z)))
          &\textit{UE}_1,\textit{EE}_1,{\to_{\mathcal{E}}}
  \end{array}\]
  The notation of the rules is taken from \cite{Comon91}. Almost all rules are reduction or
  simplification rules. The only exception is the \emph{explosion rule} \textit{E(x)} which
  performs a signature-based case distinction on the possible instantiations for the
  variable $x$: either $x=0$ or $x=s(t)$ for some term $t$.
  
  No rule is applicable to the last formula, but there is still a universal quantifier left.
  Hence the quantifier elimination is not successful.
  \qed
  \end{longitem}
\end{example}

The previous example can, alternatively, be solved using test sets~\cite{Bouhoula97,BouhoulaJouannaud97}. Test set approaches describe the minimal model of the specification by a set of rewrite rules in such a way that the query holds iff it can be reduced to a tautology (or a set thereof) by the rewrite rules. Such approaches rely on the decidability of ground reducibility~\cite{Plaisted85,Kapur91,Kounalis92,ComonJacquemard97}.

Following Bouhoula and Jouannaud, $N_G'$ corresponds to the following term re\-write system:
  \begin{align*}
    G(s(x),0)&\to\textit{true}\\
    G(x,0)&\to G(x,s(y))\\
    G(0,y)&\to\textit{false}
  \end{align*}
To prove $N_G'\not\models_{\Ind}\forall x,y.G(x,y)$, the algorithm maintains a set of currently regarded formulas with side conditions, which are all reducible to tautologies iff $N_G'\models_\Ind\forall x,y. G(x,y)$. It starts with the query $\{G(x,y)\eq \textit{true}\}$. Using the \emph{rewrite splitting} rule, a case distinction based on the possible applications of rewrite rules to $G(x,y)\eq \textit{true}$ is performed. The result is the formula set
  \[\begin{array}{rlll}
              \{&\textit{true}\eq\textit{true}&\text{ if }x=s(x')\wedge y=0&,\\
              & G(x,y')\eq\textit{true}&\text{ if }y=0\wedge y'=s(y'')&,\\
              & \textit{false}\eq\textit{true}&\text{ if }x\eq 0 &\}\ .
    \end{array}
  \]
Since the last formula is not reducible to a tautology, $N\not\models_\Ind\forall x,y. G(x,y)$ follows.

Here is a second example where all previously mentioned methods fail:

\begin{example}
  \label{ex:quantifier alternation}
  The formula $\forall x.\exists y.x\noteq 0\to G(x,y)$ is obviously valid
  in each Herbrand model of the theory
  $N_G'=\{{\to G(s(x),0)},\ {G(x,s(y))\to G(x,0)}\}$
  from Example~\ref{ex:anotherpartdef} over the signature $\Signature_{\text{nat}}=\{0,s\}$,
  i.e.~\TRonly{it holds that }%
  $N_G'\models_{\Signature_{\text{nat}}} \forall x.\exists y.x\noteq 0\to G(x,y)$.
  In our inference system, this can again be proved in a two step derivation:
  \[\begin{array}{rl@{\ \|\ }rl}
         \text{clauses in $N$:}\hfill1:&          &         &\to G(s(x),0)\\
                                     2:&          &G(x,s(y))&\to G(x,0)\\
    \text{negated conjecture: }\hfill3:&u\eq x    &x\eq 0&\to \\
                               \hfill4:&u\eq x    &G(x,y)&\to \\
       \text{Equality Resolution(3)}=5:&u\eq 0    &&\Box\\
           \text{Superposition(1,4)}=6:&u\eq s(x) &&\Box
  \end{array}\]
  The constraints $u\eq 0$ and $u\eq s(x)$ of the constrained empty clauses are covering,
  which proves that $N_G'\models_{\Signature_{\text{nat}}}\forall x.\exists y.x\noteq 0\to G(x,y)$.

  However, all previous approaches based on implicit induction formalisms fail to prove
  even the weaker proposition $N_G'\models_\Ind \forall x.\exists y.x\noteq 0\to G(x,y)$,
  because they cannot cope with the quantifier alternation.
  \qed
\end{example}

\subsection{Reasoning about $\perfect{N}$}

As we have seen in Example~\ref{ex:diverges}, a proof of $\models_\Signature$ validity using \BaseCalc\ may require the computation of infinitely many constrained empty clauses. This is not surprising, because we have to show that an existentially quantified formula cannot be satisfied by a term-generated infinite domain.
  In the context of the concrete model $\perfect{N}$ of a saturated and $\Signature$-satisfiable constrained clause set $N$, we can make use of additional structure provided by this model. To do so, we introduce a further inference that enables the termination of derivations in additional cases. The given version of this rule is in general not sound for $\models_\Signature$ but glued to the currently considered model $\perfect{N}$; however, %
analogous results hold for every Herbrand model of $N$ over $\Signature$ and even for arbitrary sets of such models, in particular for the set of all Herbrand models of $N$ over $\Signature$.

\medskip
Over any domain where an induction theorem is applicable, i.e.\ a domain on which a (non-trivial) well-founded partial ordering can be defined, we can exploit this structure to concentrate on finding minimal solutions. We do this by adding a form of induction hypothesis to the constrained clause set.
  If, e.g., $P$ is a unary predicate over the natural numbers and $n$ is the minimal number such that $P(n)$ holds, then we know that at the same time $P(n-1), P(n-2),\ldots$ do not hold.
This idea will now be cast into an inference rule (Definition~\ref{induction rule}) that can be used during a \BaseCalc-based $\models_\Ind$~theorem proving derivation (Theorem~\ref{induction rule is sound}).

\medskip
Let ${<}$ be a well-founded partial ordering on on the elements of $\perfect{N}$, i.e.~on $\Terms(\Signature)/{\stackrel *\leftrightarrow_{R_N}}$.
If $s,t$ are non-ground terms with equivalence classes $\class s$ and $\class t$, then we define $\class s< \class t$ if and only if $\class{s\sigma}< \class{t\sigma}$ for all grounding substitutions $\sigma\In X'\to\Terms(\Signature)$, where $X'\subseteq X\cup V$. The definition lifts to equivalence classes $\class\sigma,\class\rho\In X'\to\Terms(\Signature)/_{\stackrel *\leftrightarrow_R}$ of substitutions, where we say that $\class\rho<\class\sigma$ if and only if $\class{x\rho}<\class{x\sigma}$ for all $x\in X'$.

\begin{lemma}
  \label{induction rule is locally sound}
  Let $N$ be a saturated constrained clause set and let $A_N$ be not covering.
  Let $V=\{v_1,\ldots,v_k\}$, let $\alpha=v_1\eq x_1,\ldots,v_k\eq x_k$ be a constraint that
  contains only variables
  and let $X_\alpha=\{x_1,\ldots,x_k\}$ be the set of non-existential variables in $\alpha$.
  Let $H=\{\clause {C_1}\alpha,\ldots,\clause {C_n}\alpha\}$ be a set of constrained clauses containing
  only variables in $V\cup X_\alpha$.
  Furthermore, let $\rho_1,\rho_2\In X_\alpha\to\Terms(\Signature,X)$ be substitutions
  with $\class{\rho_1}<\class{\rho_2}$.
  \par
  If $N\models_{\Ind} H$, then there is a ground substitution $\sigma:V\to \Terms(\Signature)$
  such that $N\models_{\Ind}{\alpha\sigma\rho_2\to(\neg C_1\rho_1\vee\ldots\vee\neg C_n\rho_1)}$.
\end{lemma}
\begin{proof}
  Let $\class{\sigma_{\text{min}}}\In V\to\Terms(\Signature)/_{\stackrel *\leftrightarrow_R}$ be
  minimal with respect to $<$ such that
    $N\models_{\Ind}
      \{\alpha\sigma_{\text{min}}\to{C_1},
        \ldots,\alpha\sigma_{\text{min}}\to{C_n}\}$.
  We will show that $\sigma_{\text{min}}$ is the wanted substitution.
  
  Let $X_{\rho_2}$ be the set of variables in the codomain of $\rho_2$
  and let $\tau\In X_{\rho_2}\to\Terms(\Signature)$ be such that
  $N\models_{\Ind} \alpha\sigma_{\text{min}}\rho_2\tau$.
  Note that this set of ground equations equals
  $v_1\sigma_{\text{min}}\eq x_1\rho_2\tau,\ldots,v_k\sigma_{\text{min}}\eq x_k\rho_2\tau$
  because the domains of $\rho_2\tau$ and $\sigma_{\text{min}}$ are disjoint.
  We have to show that $N\models_{\Ind} \neg C_1\rho_1\tau\vee\ldots\vee\neg C_n\rho_1\tau$.
  
  To achieve a more concise representation, we employ the symbols $\forall$ and $\exists$
  on the meta level, where they are also used for higher-order quantification.
    The restriction of a substitution $\sigma$ to the set $V$ of existential variables
  is denoted by $\sigma|_V$, and $\sigma_\alpha\In V\to\Terms(X,\Signature)$ is the substitution
  induced by $\alpha$, i.e.\ $\sigma_\alpha$ maps $v_i$ to $x_i$.
  \begin{eqnarray*}
    & & \class{\rho_1} < \class{\rho_2}\\
    &\iff&
      \class{\sigma_\alpha\rho_1} < \class{\sigma_\alpha\rho_2}\\
      && \text{because $X_\alpha\subseteq X$}\\
    &%
     \implies&
      \class{(\sigma_\alpha\rho_1\tau)|_V} < \class{(\sigma_\alpha\rho_2\tau)|_V}\\
      && \text{Since $N\models_{\Ind} \alpha\sigma_{\text{min}}\rho_2\tau$,
        the latter class equals $\class{\sigma_{\text{min}}}$.}\\
    &\implies&
      N\not\models_{\Ind}
        \{\alpha(\sigma_\alpha\rho_1\tau)|_V\to C_1(\sigma_\alpha\rho_1\tau)|_V,\ldots,
        \alpha(\sigma_\alpha\rho_1\tau)|_V\to C_n(\sigma_\alpha\rho_1\tau)|_V\}\\
      && \text{because of the minimality of $\class{\sigma_{\text{min}}}$}\\
    &%
     \implies&
      \exists{\tau'}.\,
         N\models_{\Ind} \alpha(\sigma_\alpha\rho_1\tau)|_V\tau'
      \text{ and } N\not\models_{\Ind} C_1\tau'\wedge\ldots\wedge C_n\tau'\\
    &\stackrel{\phantom{f}}\implies&
      \exists{\tau'}.\, \forall i.\,
         N\models_{\Ind} {v_i\sigma_\alpha\rho_1\tau}\eq{x_i\tau'}
      \text{ and } N\not\models_{\Ind} C_1\tau'\wedge\ldots\wedge C_n\tau'\\
      &&\text{because $\tau'$ and $(\sigma_\alpha\rho_1\tau)|_V$ affect different sides
              of each equation in $\alpha$}\\
    &\stackrel{\phantom{f}}\implies&
      \exists{\tau'}.\, \forall i.\,
         N\models_{\Ind} {x_i\rho_1\tau}\eq{x_i\tau'}
      \text{ and } N\not\models_{\Ind} C_1\tau'\wedge\ldots\wedge C_n\tau'\\
    &\implies&
      \exists{\tau'}.\, \forall x\in X_\alpha.\,
         N\models_{\Ind} {x\rho_1\tau}\eq{x\tau'}
      \text{ and } N\models_{\Ind} \neg C_1\tau'\vee\ldots\vee\neg C_n\tau'\\
      && \text{because $C_1\tau'\wedge\ldots\wedge C_n\tau'$ is ground}\\
    &\implies& N\models_{\Ind} \neg C_1\rho_1\tau\vee\ldots\vee \neg C_n\rho_1\tau\\
      && \text{because $\var(C_i)\subseteq X_\alpha$}
  \end{eqnarray*}
for $i\in\{1,\ldots,k\}$ and $\tau'\In X_\alpha\to\Terms(\Signature)$.
\end{proof}

Usually when we consider sets of constrained clauses, all considered constrained clauses are supposed to have been renamed in advance so that they do not have any universal variables in common. We deviate from this habit here by forcing the common constraint $\alpha=v_1\eq x_1,\ldots,v_k\eq x_k$ upon all constrained clauses in $H$. Note that this does not affect the semantics because of the order of existential and universal quantifiers. E.g., the constrained clause set $\{\clause{P(x)\to}{u\eq x},\ \clause{\to P(y)}{u\eq y}\}$ has the semantics $\exists u.\forall x,y.(u\noteq x\vee \neg P(x))\wedge (u\noteq y\vee P(y))$, which is equivalent to the semantics $\exists u.\forall x.(u\noteq x\vee \neg P(x))\wedge (u\noteq x\vee P(x))$ of the constrained clause set $\{\clause{P(x)\to}{u\eq x},\ \clause{\to P(x)}{u\eq x}\}$.

The formula $\alpha\rho_2\to(\neg C_1\rho_1\vee\ldots\vee\neg C_n\rho_1)$ can usually not be written as a single equivalent constrained clause if some $C_i$ contains more than one literal. However, if $D_1\wedge\ldots\wedge D_m$ is a conjunctive normal form of $\neg C_1\vee\ldots\vee \neg C_n$, then each $D_j$ is a disjunction of literals and so $\clause{D_j\rho_1}{\alpha\rho_2}$ is a constrained clause.

We will now cast these ideas into an inference rule.

\begin{definition}
  \label{induction rule}
  The \emph{inductive superposition calculus \IndCalc($H$) with respect to a finite
  constrained clause set $H$} is the union of \BaseCalc\
  and the following inference rule:%
  \begin{longitem}
  \item \textbf{Induction with respect to $H$:}
  \[\inferit{\clause{C_1}\alpha\quad\ldots\quad\clause{C_n}\alpha}{\clause{D\rho_1}{\alpha\rho_2}}%
    {}\]
  where
    (i)~$H=\{\clause{C_1}\alpha,\ldots,\clause{C_n}\alpha\}$
   (ii)~$\alpha=v_1\eq x_1,\ldots,v_m\eq x_m$ \TrTocl{contains}{is a constraint containing}
        only equations between variables (and $V=\{\tenum v m\}$),
  (iii)~all variables of the premises occur in $\alpha$,
   (iv)~$\rho_1,\rho_2:\{\tenum x m\}\to \Terms(\Signature,X)$
        and $\class{\rho_1}<\class{\rho_2}$, and
    (v)~$D$ is an element of the conjunctive normal form of $\neg C_1\vee\ldots\vee\neg C_n$.
  \end{longitem}
\end{definition}

Lemma~\ref{induction rule is locally sound} ensures that all constrained clauses derived by the induction inference rule with respect to $H$ will have a common solution with the initial query $H$, because the preserved solution $[\sigma_{\text{min}}]$ is independent of the choices of $\rho_1$ and $\rho_2$.

\begin{example}
  Let $\Signature_{\text{nat}}=\{0,s\}$ and $N_P=\{P(s(s(x)))\}$.
  All clauses derivable by the induction inference rule wrt.\ $H_P=\{\clause{\to P(x)}{u\eq x}\}$
  are of one of the forms
  $\clause{P(s^n(0))\to}{u\eq s^{n+m}(0)}$, $\clause{P(s^n(x))\to}{u\eq s^{n+m}(0)}$,
  or $\clause{P(s^n(x))\to}{u\eq s^{n+m}(x)}$
  for natural numbers $n,m$ with $m>0$.
  All these formulas and the initial constrained clause set $H_P$ have
  in $\perfect{N_P}$ the common solution $\{u\mapsto s(s(0))\}$.
  \qed
\end{example}

We can thus, to decide the validity of $H$ in $\perfect{N}$, use the induction inference rule for $H$ in a theorem proving derivation:

\begin{theorem}[Soundness of the Induction Rule]
  \label{induction rule is sound}
  Let $N$ be a constrained clause set that is saturated with respect to \BaseCalc\ 
  and let $A_N$ be not covering.
  Let $V=\{v_1,\ldots,v_k\}$, let $\alpha=v_1\eq x_1,\ldots,v_k\eq x_k$ be a constraint that
  contains only variables
  and let $X_\alpha=\{x_1,\ldots,x_k\}$ be the set of non-existential variables in $\alpha$.
  Let $H$ be a finite set of constrained clauses containing only variables in $V\cup X_\alpha$.

  If $N\cup H'$ is derived from $N\cup H$ using \IndCalc($H$),
  then $N\models_{\Ind} H\iff N\models_{\Ind} H'$.
\end{theorem}
\begin{proof}
  This follows directly from Proposition~\ref{solutions do not change}, which implies that
  the solutions of $H$ are not changed by the rules in \BaseCalc,
  and Lemma~\ref{induction rule is locally sound}, which states that minimal
  solutions are invariant under the induction inference rule for $H$.
\end{proof}

This theorem basically states that the addition of constrained clauses of the presented form is a valid step in a $\models_\Ind$~theorem proving derivation that starts from $N$ and $H$ and uses the calculus \BaseCalc. Before we come to applications of the induction rule, let us shortly investigate the side conditions to this rule. Conditions (iv) and (v) are direct consequences of the ideas developed at the beginning of this section. Conditions (i)--(iii) are needed to guarantee soundness.

\begin{example}
 We present some examples to show how a violation of one of the conditions (i)--(iii) makes the induction rule unsound.
 \begin{enumerate}
 \item[(i)]
  It is important to use the induction rule on the whole query set only (condition (i)),
  because the minimal solution of a subset of the query may not be equal to the minimal solution
  of the whole query.
  Let us consider the constrained clause set $N_{\text{(i)}}=\{\to P(x),\ Q(a)\to,\ \to Q(b)\}$
  over the signature $\{a,b\}$ where $[b]<[a]$, \TrTocl{and}{and the query}
  $H_{\text{(i)}}=\{\clause{\to P(x)}{u\eq x},\ \clause{\to Q(x)}{u\eq x}\}$.
  The set $N_{\text{(i)}}\cup H_{\text{(i)}}$ is satisfiable over $\{a,b\}$: just set $u\mapsto b$.
  Using the induction rule for $H_{\text{(i)}}$, only the redundant constrained clause
  $\clause{P(a),Q(a)\to}{u\eq b}$ is derivable, namely for $\rho_1(x)=b$ and $\rho_2(x)=a$.
  If we apply the induction rule for $\{\clause{\to P(x)}{u\eq x}\}$ instead of $H_{\text{(i)}}$,
  ignoring condition (i),
  we can derive the constrained clause $\clause{P(a)\to}{u\eq b}$.
  The combined set $N_{\text{(i)}}\cup H_{\text{(i)}}\cup\{\clause{P(a)\to}{u\eq b}\}$ is
  unsatisfiable over $\{a,b\}$.
 
 \item[(ii)]
  For an example illustrating the need for condition (ii), consider the constrained clause set
  $N_{\text{(ii)}}=\{s(0)\eq 0\to,\ \to s(s(x))\eq x\}$ over the signature
  $\Signature_{\text{nat}}=\{s,0\}$.
  In the minimal model of $N_{\text{(ii)}}$, all ground terms representing even numbers
  are equivalent, as are all ground terms representing odd numbers, i.e.~there are exactly
  two equivalence classes, $[0]$ and $[s(0)]$.
  Let $[0]<[s(0)]$ and consider the query $H_{\text{(ii)}}=\{\clause{\to x\eq 0}{u\eq s(x)}\}$.
    The instantiation $u\mapsto 0$ is a witness of the validity of $H_{\text{(ii)}}$ in the
  minimal model of $N_{\text{(ii)}}$.
  However, applying the induction rule on $H_{\text{(ii)}}$ in violation of condition (ii)
  with $\rho_1(x)=0$ and $\rho_2(x)=s(0)$, we can
  derive $\clause{0\eq 0\to}{u\eq s(s(0))}$. The only instantiation validating this constrained
  clause in the minimal model of $N_{\text{(ii)}}$ is $u\mapsto 0$, i.e.~the combined set
  $H_{\text{(ii)}}\cup\{\clause{0\eq 0\to}{u\eq s(s(0))}\}$ is not valid in this model.
 
 \item[(iii)]
  \TOCLonly{\hspace{-.3em}}Now consider the \TOCLonly{empty }theory $N_{\text{(iii)}}=\{\}$ over the signature
  $\Signature_{\text{nat}}$ with $[0]<[s(0)]<[s(s(0))]<\ldots$ and the query
  $H_{\text{(iii)}}=\{\clause{y\eq x\to y\eq s(0)}{u\eq x}\}$.
  The instantiation $u\mapsto s(0)$ shows that $H_{\text{(iii)}}$ is valid in the minimal model
  $\Terms(\Signature_{\text{nat}})$ of $N_{\text{(iii)}}$.
  Note that no other instantiation of $u$ can show this.
  If we ignore condition (iii) and apply the induction rule to $H_{\text{(iii)}}$
  with $\rho_1(x)=x$ and $\rho_2(x)=s(x)$, we can derive
  $\clause{y\eq s(0)\to}{u\eq s(x)}$. This constrained clause can only be satisfied in the minimal
  model of $N_{\text{(iii)}}$ by the instantiation $u\mapsto 0$.
  Since this instantiation is not suited for $H_{\text{(iii)}}$,\TRonly{ the set}
  $H_{\text{(iii)}}\cup\{\clause{y\eq s(0)\to}{u\eq s(x)}\}$ is not valid in the minimal model
  of $N_{\text{(iii)}}$.
 \qed
 \end{enumerate}
\end{example}

Some \TRonly{further }examples will demonstrate the power of the extended calculus \IndCalc($H$).
In these examples, there will always be a unique (non-empty) set $H$ satisfying the side conditions of the induction rule, and we will write \IndCalc\ instead of \IndCalc($H$).

\TrTocl{In contrast to the other inference rules, which have a unique conclusion for each given set of premises, t}{T}he induction rule will often allow to derive an unbounded number of conclusions. So the application of this rule in all possible ways is clearly unfeasible. It seems appropriate to employ it only when a conclusion can directly be used for a superposition inference simplifying another constrained clause. We will use this heuristic in the examples below.

\begin{example}
  \label{ex:does not diverges}
  We revisit the partial definition of the usual ordering on the naturals given
  by $N_G=\{{\to G(s(0),0)},\ {G(x,y)\to G(s(x),s(y))}\}$, as shown in the introduction and in
  Example~\ref{ex:diverges}.
  Again, we want to check whether or not $N_G\models_{\Signature_{\text{nat}}}\forall x.G(s(x),x)$.
  While the derivation in Example~\ref{ex:diverges} diverges, a derivation using \IndCalc\ 
  terminates after only a few steps:
  \[\begin{array}{rl@{\ \|\ }rl}
    \text{clauses in $N_G$: }\hfill  1:&         &      &\to G(s(0),0)\\
                                     2:&         &G(x,y)&\to G(s(x),s(y))\\
    \text{negated conjecture: }\hfill3:&u\eq x   &G(s(x),x)&\to \\
           \text{Superposition(1,3)}=4:&u\eq 0   &      &\Box\\
           \text{Superposition(2,3)}=5:&u\eq s(y)&G(s(y),y)&\to \\
                 \text{Induction(3)}=6:&u\eq s(z)&      &\to G(s(z),z)\\
           \text{Superposition(6,5)}=7:&u\eq s(z)&      &\Box\\
  \end{array}\]
  The induction rule was applied using $H=\{\clause{G(s(x),x)\to}{u\eq x}\}$,
  $\rho_1(x)=z$ and $\rho_2(x)=s(z)$. At this point, the constrained clauses $\clause\Box{u\eq 0}$
  and $\clause\Box{u\eq s(z)}$ have been derived.
  Their constraints are covering for $\{s,0\}$, which means that
  $N_G\models_\Ind\forall x.G(s(x),x)$.
  Because of Proposition~\ref{models_ind = models_sig for positive universal clauses},
  this implies $N\models_{\Signature_{\text{nat}}}\forall x.G(s(x),x)$.
  \qed
\end{example}

\begin{example}
  A standard example that can be solved by various approaches
  (e.g.~\cite{GanzingerStuber92,ComonNieuwenhuis00}) is the theory
  of addition on the natural numbers:
  $N_{+}=\{\to 0+y\eq y,\ \to s(x)+y\eq s(x+y)\}$.
  A proof of $N_{+}\models_{\Ind}\forall x.x+0\eq x$ with \IndCalc\ terminates quickly:
  \[\begin{array}{rl@{\ \|\ }rl}
         \text{clauses in $N_{+}$:}\hfill1:&      &              &\to 0+y\eq y\\
                                     2:&          &              &\to s(x)+y\eq s(x+y)\\
    \text{negated conjecture: }\hfill3:&u\eq x    &x+0\eq x      &\to \\
           \text{Superposition(1,3)}=4:&u\eq 0    &0\eq 0        &\to \\
       \text{Equality Resolution(4)}=5:&u\eq 0    &              &\Box\\
           \text{Superposition(2,3)}=6:&u\eq s(y) &s(y+0)\eq s(y)&\to \\
                 \text{Induction(3)}=7:&u\eq s(z) &              &\to z+0\eq z\\
           \text{Superposition(7,6)}=8:&u\eq s(z) &s(z)\eq s(z)  &\to \\
       \text{Equality Resolution(8)}=9:&u\eq s(z) &              &\Box\\
  \end{array}\]
  The induction rule was applied using $H=\{\clause{x+0\eq x\to}{u\eq x}\}$,
  $\rho_1(x)=z$ and $\rho_2(x)=s(z)$. At this point, the constrained clauses $\clause\Box{u\eq 0}$
  and $\clause\Box{u\eq s(z)}$ have been derived.
  Their constraints cover all constraints of the form $u\eq t$,
  $t\in\Terms(\Signature_{\text{nat}},X)$, which means that
  $N_{+}\not\models_{\Ind} \clause{x+0\eq x\to}{u\eq x}$, i.e.\ 
  $N_{+}\models_{\Ind} \forall x.x+0\eq x$.
  
  Without the induction rule, the derivation in this example would resemble the one in
  Example~\ref{ex:diverges} and diverge. We would thus not even gain information about the
  $\models_{\Signature_{\text{nat}}}$ validity of the query.
  Here, however, we can again apply Proposition~\ref{models_ind = models_sig for positive universal clauses}
  to show additionally that $N_{+}\models_{\Signature_{\text{nat}}}\forall x.x+0\eq x$.
  \qed
\end{example}

Along the same lines, we can also prove that addition is symmetric, i.e.~$N_{+}\models_{\Ind}\forall x,y.x+y\eq y+x$. In this case, we need to apply the induction rule twice to obtain the additional clauses
\[\clause{\to x+y'\eq y'+x}{u\eq x,v\eq s(y')}\ \phantom{.}\]
and
\[\clause{\to x'+y\eq y+x'}{u\eq s(x'),v\eq y}\ .\]

\begin{example}
  Given the theory $N_E=\{\to E(0),\ E(x)\to E(s(s(x)))\}$ of the natural numbers together with a
  predicate describing the even numbers, we show that $N_E\not\models_{\Ind}\forall x.E(x)$.
  A possible derivation runs as follows:
  \[\begin{array}{rl@{\ \|\ }rl}
    \text{clauses in $N_E$:}\hfill   1:&             &    &\to  E(0)\\
                                     2:&             &E(x)&\to  E(s(s(x)))\\
    \text{negated conjecture: }\hfill3:&u\eq x       &E(x)&\to \\
           \text{Superposition(1,3)}=4:&u\eq 0       &    &\Box\\
           \text{Superposition(2,3)}=5:&u\eq s(s(y)) &E(y)&\to \\
                 \text{Induction(3)}=6:&u\eq s(s(z)) &    &\to E(z)\\
           \text{Superposition(6,5)}=7:&u\eq s(s(z)) &    &\Box
  \end{array}\]
  The induction rule was applied using $H=\{\clause{E(x)\to}{u\eq x}\}$,
  $\rho_1(x)=z$ and $\rho_2(x)=s(s(z))$.
  The set $\{(1)-(7)\}$ is saturated with respect to \BaseCalc.
  We could, of course, use the induction rule to derive one more non-redundant constrained clause,
  namely $\clause{\to E(z)}{u\eq s(z)}$. However, this constrained clause cannot be used in any
  further inference.
  All other constrained clauses derivable by the induction rule are redundant.
  
  The derived constrained empty clauses are $\clause{\Box}{u\eq 0}$
  and $\clause{\Box}{u\eq s(s(z))}$.
  Their constraints are not covering: They miss exactly the constraint $u\eq s(0)$,
  and in fact $N_E\models_{\Ind} E(s(0))\to$.
  
  Note that, although also $N_E\models_{\Ind} E(s(s(s(0))))\to$, we cannot derive this nor
  any other additional counterexample.
  This is due to the fact that the application of the induction rule preserves only the
  minimal satisfying constraint.
  \qed
\end{example}

\section{Conclusion} \label{secrfrc}

We have presented the superposition calculi \BaseCalc\ and \GeneralCalc, which are sound and refutationally complete for a fixed domain semantics for first-order logic. Compared to other approaches in model building over fixed domains, our approach is applicable to a larger class of clause sets.
We showed that standard first-order and fixed domain superposition-based reasoning, respectively, delivers minimal model results for some cases.
Moreover, we presented a way to prove the validity of minimal model properties by use of the calculus \IndCalc($H$), combining \BaseCalc\ and a specific induction rule.

\smallskip
The most general inductive theorem proving methods based on saturation so far are those by Ganzinger and Stuber~\cite{GanzingerStuber92} and Comon and Nieuwenhuis~\cite{ComonNieuwenhuis00}. 
Both approaches work only on sets of purely universal and universally reductive (Horn) clauses. 
Given such a clause set $N$ and a query $\forall\vec x. C$, Comon and Nieuwenhuis compute a so-called $I$-axiomatization $A$ such that $N\models_\Ind A$ and $N\cup A$ has only one Herbrand model, and then check the first-order satisfiability of $N\cup A\cup \{C\}$.
Like ours, this method is refutationally complete but not terminating. In fact, the clause set $A$ does in general not inherit properties of $N$ like universal reductiveness or being Horn, so that the saturation of $N\cup A\cup \{C\}$ does not necessarily terminate even if $N\cup\{C\}$ belongs to a finitely saturating fragment.
Ganzinger and Stuber, on the other hand, basically saturate $N\cup \{C\}$. Even if $N\cup\{C\}$ saturates finitely, this results in a non-complete procedure because productive clauses may be derived. They also present a way to guarantee completeness by forcing all potentially productive atoms to the ground level. This effectively results in an enumeration of ground instances, at the cost that the resulting algorithm almost never terminates.

We gave an example of a purely universal inductive theorem proving problem that can be solved using \BaseCalc\ while neither of the above approaches works (Example~\ref{ex:anotherpartdef}).
  Additionally, we showed how we can also prove formulas with a $\forall\exists$ quantifier alternation, i.e.\ check the validity of $\forall^*\exists^*$-quantified formulas. The opposite $\exists\forall$ quantifier alternation or subsequent alternations can currently not be tackled by our calculus and are one potential subject for future work.

Another intensely studied approach to inductive theorem proving is via test sets~\cite{Kapur91,Bouhoula97,BouhoulaJouannaud97}. Test sets rely on the existence of a set of constructor symbols that are either free or specified by unconditional equations only. Such properties are not needed for the applicability of our calculus. However, in order to effectively apply our induction rule, we need decidability of the ordering $<$ on $\perfect{N}$, i.e.~on the $\Terms(\Signature)/_{\stackrel *\leftrightarrow_R}$ equivalence classes. 
The existence of constructor symbols is often useful to establish this property. Examples~\ref{ex:diverges} and~\ref{ex:quantifier alternation} are not solvable via test sets, whereas Example~\ref{ex:anotherpartdef} is.

  Finally, works in the tradition of Caferra and Zabel~\cite{CaferraZabel92} or Kapur~\cite{Kapur91,KapurSubramaniam00,GieslKapur03,FalkeKapur06} consider only restricted forms of equality literals and related publications by Peltier~\cite{Peltier03} pose strong restrictions on the clause sets (e.g.\ that they have a unique Herbrand model).

In summary, our approach does not need many of the prerequisites required by previous approaches, 
like solely universally reductive clauses in $N$, solely Horn clauses, solely purely universal clauses, solely non-equational clauses, the existence and computability of an ``$A$'' set making the minimal model the unique Herbrand model, or the existence of explicit constructor symbols. Its success is built on a superposition-based saturation concept.

\smallskip
There are several obvious ways in which to extend the presented calculi.
In analogy to the work of Bachmair and Ganzinger~\cite{BachmairGanzinger94}, it is possible 
to extend the new superposition calculi by negative literal selection, with the restriction that no constraint literals may be selected. This does not affect refutational completeness.
  For universally reductive clause sets $N$, it is also possible to make the inductive theorem proving calculus \IndCalc($H$) (with selection) refutationally complete, following the approach of Ganzinger and Stuber~\cite{GanzingerStuber92}.
As in their context, this particular superposition strategy carries the disadvantage of enumerating all ground instances of all clauses over to our setting. So it can hardly be turned into a decision procedure for clause classes having infinite Herbrand models.
  In some cases, the induction rule might constitute a remedy: In case we can finitely saturate a clause set $N$, the ordering $<$ on its minimal model $\perfect{N}$ may become effective and hence the induction rule may be effectively usable to finitely saturate clause sets that otherwise have an infinite saturation.

Our hope is that the success of the superposition-based saturation approach on identifying decidable classes with respect to the classical first-order semantics can be extended to some new classes for the fixed domain and minimal model semantics. Decidability results for the fixed domain semantics are hard to obtain for infinite Herbrand domains but the problem can now be attacked using the sound and refutationally complete calculus \BaseCalc. This will require in addition the extension of the redundancy notion suggested in Section~\ref{secforifx} as well as more expressive constraint languages. Here, concepts and results from tree automata could play a role. First results in this direction have been established~\cite{HorbachWeidenbach09CSL}.

It also turns out that an extension of the current algorithm can be employed to decide the validity of various classes of formulae with a $\forall^*\exists^*$ or $\exists^*\forall^*$ prefix in models that are represented by a conjunction of atoms or by the contexts computed in model evolution~\cite{HorbachWeidenbach09CADE}.

\begin{acks}
We want to thank our reviewers for their valuable and detailed comments. Their suggestions helped significantly in improving this article.

Matthias Horbach and Christoph Weidenbach are supported by the German Trans\-regional Collaborative Research Center SFB/TR~14 AVACS.
\end{acks}

\begin{received}
Received September 2008;
revised July 2009;
accepted October 2009
\end{received}

\end{document}